\documentclass[conference]{IEEEtran}
\usepackage{tikz}
\usepackage{amsmath,amsfonts,amssymb,graphicx}
\usepackage{wasysym}
\usepackage{footnote}
\usepackage{multirow}
\usepackage{pifont}
\usepackage{subfigure}
\usepackage[lined,ruled,commentsnumbered]{algorithm2e}
\usepackage{algpseudocode}
\usepackage{pdfpages}
\usepackage{tabularx}
\usepackage{booktabs}
\usepackage[font={small}]{caption}
\usepackage{tcolorbox}
\usepackage{hyperref}
\usepackage{makecell}
\usepackage{float}
\usepackage{xcolor}
\usepackage{enumitem}

\begin{document}
%
\title{CAT: Can Trust be Predicted with Context-\\Awareness in Dynamic Heterogeneous Networks?}

	

%
\author{
\IEEEauthorblockN{Jie Wang\IEEEauthorrefmark{1},
Zheng Yan\IEEEauthorrefmark{1}\IEEEauthorrefmark{2}\textsuperscript{\ding{41}},
Jiahe Lan\IEEEauthorrefmark{1}, 
Xuyan Li\IEEEauthorrefmark{2}, and
Elisa Bertino\IEEEauthorrefmark{3}}
\IEEEauthorblockA{\IEEEauthorrefmark{1}State Key Laboratory of Integrated Services Networks, School of Cyber Engineering, Xidian University}
\IEEEauthorblockA{\IEEEauthorrefmark{2}Hangzhou Institute of Technology, Xidian University \ \IEEEauthorrefmark{3}Department of Computer Science, Purdue University}
\IEEEauthorblockA{jwang1997@stu.xidian.edu.cn, zyan@xidian.edu.cn, jhlan16@stu.xidian.edu.cn, xli77144@gmail.com, bertino@purdue.edu}
}

\IEEEoverridecommandlockouts
\makeatletter\def\@IEEEpubidpullup{6.5\baselineskip}\makeatother
\IEEEpubid{\parbox{\columnwidth}{
    Network and Distributed System Security (NDSS) Symposium 2026\\
    23-27 February 2026, San Diego, CA, USA\\
    ISBN 979-8-9919276-8-0\\
    https://dx.doi.org/10.14722/ndss.2026.242171\\
    www.ndss-symposium.org
}
\hspace{\columnsep}\makebox[\columnwidth]{}}

\maketitle

\begin{abstract}
Trust prediction provides valuable support for decision-making, risk mitigation, and system security enhancement. Recently, Graph Neural Networks (GNNs) have emerged as a promising approach for trust prediction, owing to their ability to learn expressive node representations that capture intricate trust relationships within a network. However, current GNN-based trust prediction models face several limitations: (i) Most of them fail to capture trust dynamicity, leading to questionable inferences. (ii) They rarely consider the heterogeneous nature of real-world networks, resulting in a loss of rich semantics. (iii) None of them support context-awareness, a basic property of trust, making prediction results coarse-grained.

To this end, we propose CAT, the first \underline{C}ontext-\underline{A}ware GNN-based \underline{T}rust prediction model that supports trust dynamicity and accurately represents real-world heterogeneity. CAT consists of a graph construction layer, an embedding layer, a heterogeneous attention layer, and a prediction layer. It handles dynamic graphs using continuous-time representations and captures temporal information through a time encoding function. To model graph heterogeneity and leverage semantic information, CAT employs a dual attention mechanism that identifies the importance of different node types and nodes within each type. For context-awareness, we introduce a new notion of meta-paths to extract contextual features. By constructing context embeddings and integrating a context-aware aggregator, CAT can predict both context-aware trust and overall trust. Extensive experiments on three real-world datasets demonstrate that CAT outperforms five groups of baselines in trust prediction, while exhibiting strong scalability to large-scale graphs and robustness against both trust-oriented and GNN-oriented attacks.
\end{abstract}


%
\IEEEpeerreviewmaketitle

\section{Introduction} \label{Introduction}
Trust is a complex and multifaceted concept, referring to a subjective view held by one entity towards another within a specific context. It is characterized by subjectivity, dynamicity, context-awareness, asymmetry, and conditional transitivity~\cite{sherchan2013survey}. Trust evaluation, which quantifies trust by considering factors affecting it, is one of the important approaches for achieving cyber trust. It has been applied to various fields for such purposes as fraud detection, intrusion detection, service management, and access control~\cite{trustguard,wang_reputation}. For example, social networks are often affected by fraudsters and dishonest ratings, as studies have reported that 10\% of Twitter users are fake~\cite{wang2017sybilscar} and that 16\% of restaurant ratings on Yelp are dishonest~\cite{liu2017efficiently}. In such cases, trust evaluation helps identify trustworthy entities and filter out misleading information, thereby improving the trustworthiness of the network. Overall, trust evaluation plays a crucial role in facilitating decision-making, mitigating risks, and enhancing system security.

\newcommand{\minitab}[2][l]{\begin{tabular}{#1}#2\end{tabular}}
\begin{table*}[htbp]
\footnotesize
\centering
\caption{Comparison between CAT and existing GNN-based trust prediction models.}
\label{related_work_table}
{\vspace{-1mm}\CIRCLE: it satisfies a criterion; \Circle: it does not satisfy a criterion; \LEFTcircle: it partially satisfies a criterion.}
\\[2mm]

\begin{tabular}{c|ccccccc|c}
\toprule[1.5pt]
\multicolumn{1}{c|}{\begin{tabular}[c]{@{}c@{}}\textbf{Models} \end{tabular}} & 
\multicolumn{1}{c}{\begin{tabular}[c]{@{}c@{}}Guardian~\cite{lin2020guardian}\end{tabular}}&
\multicolumn{1}{c}{\begin{tabular}[c]{@{}c@{}}Medley~\cite{lin2021medley}\end{tabular}}& 
\multicolumn{1}{c}{\begin{tabular}[c]{@{}c@{}}GATrust~\cite{jiang2022gatrust}\end{tabular}}&
\multicolumn{1}{c}{\begin{tabular}[c]{@{}c@{}}TrustGNN~\cite{huo2023trustgnn}\end{tabular}}&
\multicolumn{1}{c}{\begin{tabular}[c]{@{}c@{}}KGTrust~\cite{KGTrust}\end{tabular}}&
\multicolumn{1}{c}{\begin{tabular}[c]{@{}c@{}}DTrust~\cite{DTrust}\end{tabular}}& 
\multicolumn{1}{c}{\begin{tabular}[c]{@{}c@{}}TrustGuard~\cite{trustguard}\end{tabular}} &
\multicolumn{1}{|c}{\begin{tabular}[c]{@{}c@{}}\textbf{CAT} \end{tabular}} \\
\midrule
Dynamicity & \Circle & \CIRCLE & \Circle & \Circle & \Circle & \CIRCLE & \CIRCLE & \CIRCLE \\
Heterogeneity & \Circle & \Circle & \Circle & \Circle & \CIRCLE & \Circle & \Circle & \CIRCLE \\
Context-Awareness & \Circle & \Circle & \Circle & \Circle & \Circle & \Circle & \Circle & \CIRCLE \\
Robustness & \Circle & \Circle & \Circle & \Circle & \Circle & \Circle & \LEFTcircle & \CIRCLE \\

\bottomrule[1.5pt]
\end{tabular}

\vspace{-2mm}
\end{table*}

Graph Neural Networks (GNNs), an emerging Machine Learning (ML) paradigm designed for graph-structured data, have demonstrated strong performance in various tasks, such as anomaly detection~\cite{yang2023prographer,rehman2024flash,hsieh2024netvigil}, social recommendation~\cite{gnn_social,yang2021consisrec}, and malware analysis~\cite{chen2023guided,yan2019classifying}. Their strength lies in the ability to learn expressive node representations (i.e., embeddings) through information propagation and aggregation, fully leveraging the structural information inherent in graphs. This capability has motivated researchers to apply GNNs to trust evaluation, also known as trust prediction~\footnote{We use the terms ``trust evaluation'' and ``trust prediction'' interchangeably since GNNs focus on predicting future trust relationships based on historical interactions.}. Compared with statistical~\cite{chen2016trust}, reasoning~\cite{liu2019trust}, and traditional ML approaches~\cite{neuralwalk}, using GNNs for trust prediction offers several distinct advantages. First, trust relationships and other interactions between nodes can be naturally modeled as graphs, while GNNs are well-suited for processing graph data. Second, the message-passing mechanism of GNNs is compatible with basic trust properties, including conditional transitivity and composability, thus enhancing prediction accuracy. Third, unlike traditional ML approaches that require heavy and complex feature engineering, GNNs provide an end-to-end prediction mode, greatly simplifying the process of trust prediction. Consequently, GNNs present a promising approach for intelligent and precise trust prediction.

\begin{figure}
	\centering
	\includegraphics[scale=0.45]{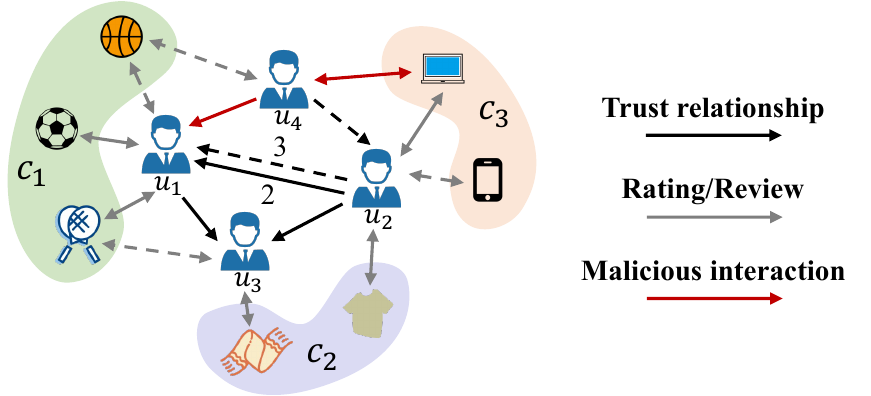}
    \caption{A motivating example of dynamic heterogeneous networks represented by a real-world social network with interactions involving users $u_1 \sim u_4$ and items across various contexts $c_1 \sim c_3$, including sports, electronics, and clothing. Dashed and solid arrows indicate interactions occurring at different time points.}
	\label{example}
    \vspace{-2mm}
\end{figure}

\textbf{Existing Approaches and Motivations.}
Table~\ref{related_work_table} summarizes existing GNN-based trust prediction models in terms of dynamicity, heterogeneity, context-awareness, and robustness. We use social networks as an example to illustrate the practical significance of these properties, as shown in Fig.~\ref{example}. Based on our review, we make the following observations: 

(i) Most models~\cite{lin2020guardian,jiang2022gatrust,huo2023trustgnn,KGTrust} do not account for trust dynamicity, resulting in questionable inferences and poor prediction accuracy~\cite{khoury2023jbeil,han2020unicorn}. As illustrated in Fig.~\ref{example}, user interactions evolve over time, and trust relationships often change accordingly; for example, $u_2$ has different trust levels towards $u_1$ at different time points. Static models fail to capture such dynamics and may mistakenly utilize later interactions to predict early trust relationships, thereby violating temporal causality.

(ii) Almost all models~\cite{trustguard,lin2020guardian,lin2021medley,jiang2022gatrust,huo2023trustgnn,DTrust} overlook the heterogeneous nature of real-world networks. Fig.~\ref{example} presents a typical heterogeneous network with various types of nodes (i.e., users and items) and edges (i.e., trust relationships between users and user ratings on items). Most trust prediction models focus solely on homogeneous user-to-user networks and thus miss valuable semantic information, such as user preferences or item characteristics, which have been shown to enhance trust prediction~\cite{KGTrust,cdeeptrust}.

(iii) None of the existing models support context-awareness, which is essential for several reasons. First, trust is inherently context-aware, as a study~\cite{tang2012mtrust} found that fewer than 1\% of people trust their friends in all contexts in real-world networks. Here, \textit{context} refers to any information that describes the specific situations in which a trust relationship is established, such as item categories (as illustrated in Fig.~\ref{example}), interaction domains, or task types. For example, $u_2$ may trust $u_1$ in sport ($c_1$) recommendations due to $u_1$'s expertise, but not in clothing ($c_2$) recommendations. Second, since different contexts influence trust in distinct ways, modeling trust within each context can improve overall trust prediction. Third, evaluating trust in one context helps predict trust in similar yet unobserved contexts. Last, context-aware models enable flexible trust-based applications by offering fine-grained evaluation results. For instance, one can identify malicious behaviors of others in a specific context based on context-aware trust~\footnote{A context-aware trust level reflects a trustor's trust in a trustee within a specific context, while an overall trust level aggregates these context-specific levels to provide a holistic view of the trustor's general trust in the trustee.}.

(iv) Resilience to data poisoning attacks is rarely evaluated in existing models~\cite{lin2020guardian,lin2021medley,jiang2022gatrust,huo2023trustgnn,KGTrust,DTrust}, hindering their practical adoption. While TrustGuard~\cite{trustguard} addresses trust-oriented attacks that operate at the node level, it overlooks more sophisticated GNN-oriented attacks that operate at the graph-structural level. In practice, attackers can inject malicious interactions (red lines in Fig.~\ref{example}) into training data to mislead model learning. Such attacks are prevalent in real-world systems; for example, Taobao has recognized service boosting via malicious ratings as a threat to its platform~\cite{liu2017efficiently}.

\textbf{Therefore, it is critical to develop a trust prediction model that can capture trust dynamicity, represent real-world situations, support context-awareness, and be resilient to data poisoning attacks.}

\textbf{Technical Challenges.}
However, designing such a model is non-trivial as it requires addressing the following challenges:

\textbf{TC1:} Modeling dynamicity with fine granularity while maintaining scalability is a challenging task. 
Time is a continuous variable that requires precise representation and careful handling. Some trust prediction models~\cite{trustguard,DTrust} use discrete-time methods to represent dynamic trust networks. While efficient, they may miss critical temporal patterns across snapshots. Continuous-time methods~\cite{lin2021medley} address this issue by considering all timestamps but suffer from low scalability. Therefore, how to achieve both accurate dynamicity modeling and scalability calls for an effective solution.

\textbf{TC2:} It is difficult to extract key information from heterogeneous graphs.
Real-world networks often contain multiple types of nodes and interactions~\cite{SiamHAN}, which provide rich semantics for accurate trust prediction. However, this diversity may also introduce excessive noise that undermines trust propagation and aggregation. Thus, how to determine the proper importance of interactions formed by different types of nodes requires a thorough investigation.

\textbf{TC3:} The absence of labeled, context-specific trust data poses a significant challenge for predicting trust across varied contexts. Existing datasets only provide labels for overall trust, making direct context-aware trust prediction infeasible. Moreover, in graph structures, contexts are less obvious than nodes and edges. Hence, how to incorporate contextual information (\textbf{TC3-1}) and predict context-aware trust from available overall trust data (\textbf{TC3-2}) needs ingenious designs.

\textbf{Our Proposal.}
In this paper, we propose CAT, the first GNN-based trust prediction model that simultaneously captures trust dynamicity, handles real-world network heterogeneity, supports context-awareness, and is resilient to data poisoning attacks. CAT adopts a layered architecture that includes a graph construction layer, an embedding layer, a heterogeneous attention layer, and a prediction layer. The graph construction layer builds a dynamic heterogeneous graph using timestamped interactions, enabling fine-grained modeling of dynamicity. The embedding layer initializes embeddings for nodes, time, and edge attributes. To address \textbf{TC3-1}, we introduce a new notion: context-aware meta-path, which incorporates contextual information while enhancing semantic representation. Meanwhile, a time encoding function maps the continuous-time domain into a vector space. The heterogeneous attention layer propagates and aggregates trust information to refine embeddings. It employs a dual attention mechanism to learn the importance of node types and the nodes within each type, addressing \textbf{TC2}. To resolve the contradiction in \textbf{TC1}, we adopt recent-time neighbor sampling and one-hop trust propagation strategies, focusing on limited yet crucial interactions. Last, the prediction layer predicts the trust relationship between any two users under a specific context. To tackle \textbf{TC3-2}, we treat item categories as contexts and create context embeddings by averaging the embeddings of items within the same context. This enables CAT to predict context-aware trust. To avoid dependency on context-specific trust labels, we further propose a context-aware aggregator that links context-aware trust with overall trust. By jointly addressing \textbf{TC2} and \textbf{TC3}, CAT gains a robust semantic understanding, forcing attackers to consider multiple factors for a successful attack.

\textbf{Evaluation.}
We conduct extensive experiments to evaluate CAT's effectiveness, scalability, and robustness on three real-world social network datasets: Epinions~\cite{tang2015trust}, Ciao~\cite{tang2015trust}, and CiaoDVD~\cite{guo2014etaf}. Experimental results show that CAT outperforms five groups of baselines, especially in predicting trust for unobserved users that do not have interactions at the training stage, achieving a 50.79\% Mean Reciprocal Rank (MRR) improvement on Epinions over the best baseline. Notably, CAT can predict context-aware trust, which is not supported by all baseline models. It also scales well to large graphs, reducing average running time by 73.97\% on Epinions. Furthermore, CAT shows strong robustness, with maximum performance drops of only 0.95\% and 3.39\% under trust-oriented~\cite{trustguard} and GNN-oriented attacks~\cite{lee2024spear}, respectively, when predicting trust for observed users.

\textbf{Contributions.} 
The main contributions are as follows:
\begin{itemize}
\item We propose CAT, the first context-aware GNN-based trust prediction model, which also uniquely addresses the dynamic nature of trust in heterogeneous networks.

\item We introduce the notion of context-aware meta-paths that incorporate contextual features and generalize across domains. By constructing context embeddings, CAT can predict context-aware trust. A context-aware aggregator is further proposed to avoid reliance on context-specific trust labels and enable CAT to predict overall trust.

\item We model trust dynamicity using a time encoding function and extract key semantic information through a dual attention mechanism. To enhance scalability, we further adopt recent-time neighbor sampling and one-hop trust propagation strategies.

\item We conduct a comprehensive evaluation on three real-world datasets. The results show that CAT effectively predicts trust for both observed and unobserved users, scales well to large graphs, and remains robust against both trust-oriented and GNN-oriented attacks.

\end{itemize}

\section{Related Work}
This section briefly reviews recent advances in dynamic heterogeneous GNNs and GNN-based trust prediction.

\subsection{Dynamic Heterogeneous GNNs} \label{dy_het_gnn}
Many research efforts have focused on either dynamic GNNs~\cite{tgn} or heterogeneous GNNs~\cite{han}. However, research on integrating both dynamicity and heterogeneity is relatively limited~\cite{zhang2023dynamic}. DyHATR~\cite{xue2021modeling} employs node-level and edge-level attentions to capture heterogeneous information, and applies Recurrent Neural Networks (RNNs) with a self-attention mechanism to discern temporal patterns. HTGNN~\cite{fan2022heterogeneous} adopts a hierarchical aggregation mechanism to consider both heterogeneous spatial and temporal features. While DyHATR and HTGNN support dynamicity by modeling a sequence of static snapshots, they overlook the fine granularity of time. In contrast, HGT~\cite{hgt} incorporates fine-grained timestamps using a relative temporal encoding technique. Additionally, it leverages meta-relations with Transformer-based attention~\cite{vaswani2017attention} to handle graph heterogeneity. However, HGT requires stacking multiple GNN layers to capture high-order information, leading to over-smoothing and poor scalability~\cite{liu2023meta}.

\textbf{Discussion.} Although these models achieve impressive performance across several tasks, their effectiveness in trust prediction remains uncertain. Specifically, basic trust properties (e.g., context-awareness and asymmetry) and trust-related attributes (e.g., trust levels and ratings) are not explicitly modeled, which may compromise prediction accuracy. Moreover, the joint modeling of dynamicity and heterogeneity often raises scalability concerns. Therefore, how to adapt existing dynamic heterogeneous GNNs to trust prediction while ensuring scalability requires further extensive investigation.

\subsection{GNN-based Trust Prediction}
Generally, GNN-based trust prediction models can be classified into static models and dynamic models, depending on whether they incorporate temporal information.

\textbf{Static Models.}
Lin \textit{et al.}~\cite{lin2020guardian} proposed Guardian, the first GNN-based trust prediction model for online social networks. Guardian leverages the concepts of in-degree and out-degree to model trust asymmetry, and utilizes localized graph convolutions for trust propagation. GATrust~\cite{jiang2022gatrust} improves Guardian by fusing multi-faceted user properties, including user features, known trust relationships, and network structure, using learnable weights. Unlike the above two models, TrustGNN~\cite{huo2023trustgnn} models trust by specifying different trust chains and then applies an attention mechanism to discriminate their importance. However, the above models focus on homogeneous user-to-user trust networks, ignoring the heterogeneous interactions between different node types found in real-world networks. KGTrust~\cite{KGTrust} addresses this issue by initializing embeddings with type-specific semantics and employing a discriminative convolutional mechanism for embedding learning. Despite these advances, all these models operate on static snapshots and thus fail to capture the dynamic nature of trust. This oversight leads to information loss and questionable inferences~\cite{khoury2023jbeil,han2020unicorn}, as they miss the sequence in which trust relationships are established. Moreover, the lack of support for context-awareness and robustness further limits their effectiveness in practice.

\textbf{Dynamic Models.}
Current dynamic models adopt two approaches for modeling time: discrete-time and continuous-time. DTrust~\cite{DTrust} and TrustGuard~\cite{trustguard} use the discrete-time method that treats a dynamic graph as a series of time-ordered snapshots. They first learn static spatial features within a snapshot, and then capture temporal patterns across snapshots through Gated Recurrent Units (GRUs) and a self-attention mechanism, respectively. While efficient, this method may lose substantial structural dependencies across snapshots~\cite{hgt}. In addition, DTrust lacks robustness considerations, whereas TrustGuard employs a robust aggregator to counter trust-oriented attacks based on the network theory of homophily. However, TrustGuard's efficacy under GNN-oriented attacks remains unclear. Medley~\cite{lin2021medley} is a continuous-time method that fully utilizes each timestamp through a time encoding function. It achieves high prediction accuracy but suffers from expensive costs and lacks robustness. Moreover, all these models do not support heterogeneity and context-awareness.

\textbf{Discussion.}
Despite remarkable progress in GNN-based trust prediction, several critical issues remain unresolved, as shown in Table~\ref{related_work_table}. First, insufficient attention is given to trust dynamicity modeling, particularly the efficient encoding of fine-grained timestamps. Second, heterogeneity is not well addressed in existing work. Third, none of the models support context-awareness, a basic nature of trust. Last, the resilience of these models to both trust-oriented and GNN-oriented attacks has been largely overlooked.
\section{Problem Statement}
In this section, we first introduce key definitions and formulate the problem we aim to address. Then, we describe the threat model of CAT.

\subsection{Problem Definition} \label{problem_definition}
\noindent \textbf{Definition 1. Heterogeneous Network.} 
A heterogeneous network, also known as Heterogeneous Information Network (HIN), is represented as a six-tuple $\mathcal{G}=(\mathcal{V},\mathcal{E}, \mathcal{T}, \mathcal{R}, \Phi, \Psi)$, where each node $v \in \mathcal{V}$ and each edge $e \in \mathcal{E}$ are associated with their type mapping functions $\Phi:\mathcal{V}\rightarrow \mathcal{T}$ and $\Psi: \mathcal{E}\rightarrow \mathcal{R}$, respectively. A network is considered heterogeneous when $|\mathcal{T}|+|\mathcal{R}|>2$, while it is homogeneous when $|\mathcal{T}|=|\mathcal{R}|=1$. 

\noindent \textbf{Definition 2. Dynamic Heterogeneous Network.}
A dynamic heterogeneous network observed within a time interval $[0,T]$ can be expressed as $\mathcal{G}(T)=(\mathcal{V}(T),\mathcal{E}(T), \mathcal{T}, \mathcal{R}, \Phi, \Psi)$, where $\mathcal{V}(T)$ and $\mathcal{E}(T)$ are time-varying sets of nodes and edges, respectively. $\mathcal{E}(T) = \{ (e_{i,j}(t), r) \mid v_i(t), v_j(t) \in \mathcal{V}(T), r \in \mathcal{R}, t \leq T \}$, where $e_{i,j}(t)$ is an edge between nodes $v_i$ and $v_j$ built at timestamp $t$, and $r$ is the edge type.

\noindent \textbf{Definition 3. Meta-path.} A meta-path $p$ of length $s$ is defined as a path in the form of $\tau_1 \xrightarrow{r_1} \tau_2 \xrightarrow{r_2} \cdots \xrightarrow{r_s} \tau_{s+1}$ with node types $\tau_1, \tau_2, \cdots, \tau_{s+1} \in \mathcal{T}$ and edge types $r_1, r_2, \cdots, r_{s} \in \mathcal{R}$. Note that any two node types and any two edge types in this path can be the same. The meta-path $p$ describes a composite relation $r_1 \circ r_2 \circ \cdots \circ r_s$ between node types $\tau_1$ and $\tau_{s+1}$, capturing rich semantics. 

Take a heterogeneous social network (Fig.~\ref{example}) as an example, the meta-path $user \xrightarrow{rate} item \xrightarrow{rate^{-}} user$, where $rate^{-}$ means ``is rated by'', describes the relationship ``two users have rated the same item'', indicating similar preferences or potential trust between them.

\noindent \textbf{Definition 4. Contextual Trust Network.}
A contextual trust network is a dynamic heterogeneous network with contexts involved, denoted as $\mathcal{G}(T)=(\mathcal{V}(T),\mathcal{E}(T), \mathcal{T}, \mathcal{R}, \Phi, \Psi, \mathcal{C})$. Here, $\mathcal{T}=\{user,item\}$, $\mathcal{R}=\{\langle user, user \rangle, \langle user, item \rangle, \langle item, user \rangle\}$, and $\mathcal{C}$ denotes the set of contexts.

In Definition~4, the term $user$ refers to either a trustor or a trustee --- two entities involved in a trust relationship, while the term $item$ has application-specific meanings, such as products in social networks or assets in financial networks. Based on the datasets employed in this paper, each context $c \in \mathcal{C}$ corresponds to an item category. Among the edge types, $ \langle user, user \rangle$ represents a directed trust relationship and is inherently asymmetric, while $\langle user, item \rangle$ and $\langle item, user \rangle$ represent inverse relationships, such as ``rate'' and ``is rated by''. The absence of $\langle item, item \rangle$ relationships is based on the assumption that the primary interactions of interest occur between $users$ and $items$ or between $users$ themselves. Additionally, the set of context-aware trust relationships is represented as $\{ (\langle v_i, v_j \rangle, c_k, t, w_{i,j,k}, r) \mid e_{i,j}(t) \in \mathcal{E}(T), c_k \in \mathcal{C}, r=\langle user, user \rangle \}$, where $w_{i,j,k}$ denotes the trust level of the trustor-trustee pair $\langle v_i, v_j \rangle$ under context $c_k$. The value of $w_{i,j,k}$ can be binary or multi-level depending on specific scenarios. Notably, due to the asymmetric nature of trust, $w_{i,j,k}$ is not necessarily equal to $w_{j,i,k}$.

\noindent \textbf{Definition 5. Trust Prediction.}
Given a contextual trust network $\mathcal{G}(T)$ observed in the time interval $[0,T]$, the goal of trust prediction is to learn a model $f(\cdot)$ that can predict, at time $T+\Delta(T)$, both the context-aware trust level $\bar{w}_{i,j,k}$ for any trustor-trustee pair $\langle v_i, v_j \rangle$ within a specific context $c_{k}$ and the overall trust level $\bar{w}_{i,j}$ (aggregated across all contexts) of this pair. Here, $v_i, v_j \in \mathcal{V}(T)$ represent user nodes, $v_i \ne v_j$, and $c_k \in \mathcal{C}$.

The trust prediction task may vary depending on the type of available supervision. If only labels for trusted relationships are provided, with distrusted ones implicitly represented by the absence of links, the task reduces to link prediction: predicting whether a trusted relationship (link) exists between two users. Conversely, if labels for trust strength (e.g., numerical scores) are available, the task becomes edge classification: predicting the trust level (edge weight) between users.

\subsection{Threat Model} \label{threat_model}
An important requirement for designing a GNN-based trust prediction model is that the model can withstand both trust-oriented~\cite{trustguard} and GNN-oriented attacks~\cite{lee2024spear}. Trust-oriented attacks operate at the local, node level, targeting individual trustworthiness by manipulating specific trust relationships. For example, attackers may destroy or boost the trustworthiness of certain nodes by launching bad-mouthing or good-mouthing attacks. In contrast, GNN-oriented attacks typically operate at the global, graph-structural level, exploiting inherent vulnerabilities of GNNs and highlighting the adversarial relationship between the attacker and the model. In this case, attackers may have access to the full input graph and strategically manipulate its structural patterns. Both types of attacks are prevalent in real-world scenarios~\cite{liu2017efficiently,lee2024spear} and usually manifest as data poisoning attacks, where attackers degrade model performance by manipulating links (e.g., trust relationships) during training. Therefore, this paper analyzes model robustness under both attacks.

We make the following assumptions. For trust-oriented attacks, we follow the approach proposed in~\cite{trustguard} and focus on collaborative bad-mouthing attacks (more disruptive than good-mouthing attacks), where a group of malicious nodes collude to destroy the trustworthiness of target nodes. For GNN-oriented attacks, we consider a powerful attacker who can access and analyze the contextual trust network $\mathcal{G}(T)$, including its statistics and temporal evolution patterns. The attacker can manipulate the network structure by strategically adding or removing links, e.g., by creating a surrogate model to guide the attack. However, the attacker lacks knowledge of the architecture or internal parameters of the trust prediction model and cannot query it. These settings are realistic and consistent with existing work~\cite{lee2024spear}. In practice, attackers typically do not know the details of a victim model, yet they can observe the input data (e.g., public social networks) and formulate attack strategies using surrogate models. We further assume that during the test phase, the input to the trust prediction model is a pair of nodes indicating the roles of trustor and trustee. Thus, evasion attacks are not applicable. Instead, data poisoning attacks may occur during the training phase to degrade the model's overall performance.

Beyond these two attacks, we also consider an adaptive data poisoning attack that specifically targets a context-aware trust prediction model by manipulating interactions across contexts to blur contextual boundaries (see Section~\ref{discussion}).

\begin{figure*}[htbp]
	\centering
	\includegraphics[scale=0.32]{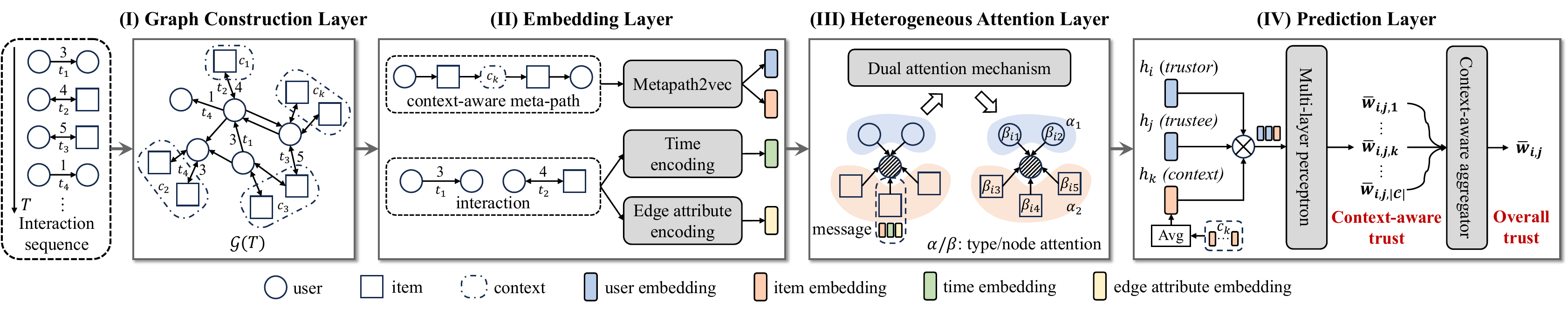}
        \vspace{-4mm}
	\caption{CAT overview.}
	\label{cat_design}
        \vspace{-3mm}
\end{figure*}
\section{CAT Design}
In this section, we first provide an overview of CAT and then elaborate on its technical details.

\subsection{Overview}
As illustrated in Fig.~\ref{cat_design}, CAT consists of four layers: a graph construction layer, an embedding layer, a heterogeneous attention layer, and a prediction layer. Specifically, the graph construction layer builds a contextual trust graph using streaming user-item and user-user interactions. In the embedding layer, a meta-path covering rich semantics and contextual features is first defined, and then node embeddings are initialized via Metapath2vec~\cite{dong2017metapath2vec}. Meanwhile, time information and edge attributes (e.g., ratings from users on items) are encoded to facilitate information propagation and aggregation. The heterogeneous attention layer employs a dual attention mechanism to discriminate the importance of node types and the importance of nodes within the same type. Leveraging the message-passing mechanism of GNNs, CAT achieves trust propagation with varying attentions granted on different interactions. To enhance scalability, we adopt recent-time neighbor sampling and one-hop trust propagation strategies, focusing on limited yet crucial interactions. Finally, in the prediction layer, we first generate a context embedding by averaging item embeddings within the same context. Then, two user embeddings together with the context embedding are fed into a Multi-Layer Perceptron (MLP) to predict the latent trust level between the two users within that context. To overcome the lack of context-specific trust labels, we propose a context-aware aggregator to link context-aware trust with overall trust, weighted by context importance. The integration of these layers gives CAT a comprehensive semantic understanding, which hinders potential attacks, as attackers have to consider multiple contextual and semantic factors. 

\subsection{Graph Construction Layer}
This layer aims to construct a contextual trust graph using user-user and user-item interactions, each associated with attributes such as timestamps, trust levels, and ratings. To effectively model the temporal dynamics of these interactions, we employ continuous-time representations, which describe the evolution of the graph as a function of continuous time. Unlike discrete-time representations~\cite{trustguard,DTrust,euler} that divide time into fixed intervals and may lose fine-grained temporal patterns, continuous-time representations allow interactions and updates to occur at arbitrary time points, enabling a more accurate and realistic modeling of trust dynamics. In addition, items belonging to the same category are grouped under the same context.

\subsection{Embedding Layer} \label{embedding_layer}
This layer aims to initialize embeddings for nodes, time, and edge attributes.

\textbf{Node Embedding.}
Typically, we can apply two transformation matrices for users and items to characterize their heterogeneity. While this method is straightforward, it fails to capture rich semantics embedded in diverse relations. Meta-paths have emerged as a promising solution to this limitation. However, conventional meta-paths, which focus on explicit links, lack the capacity to incorporate contextual information, a crucial factor for context-aware trust prediction. To address this, we extend the concept of meta-paths by introducing a new context-aware meta-path, defined as: $user \xrightarrow{rate} item \xrightarrow{belong \, to} context \xrightarrow{belong \, to^{-}} item \xrightarrow{rate^{-}} user$, simplified as $uiciu$. It is important to clarify that the ``context'' herein is not an actual node type but rather a latent factor that reflects item characteristics. This meta-path describes the relationship ``two users have interacted with two items belonging to the same context'', implying that these two users share similar preferences in that context. Then, we apply the Metapath2vec model~\cite{dong2017metapath2vec} to generate semantically-rich embeddings $h_i,h_j \in \mathbb{R}^{d_v}$, where $d_v$ is the embedding dimension. 
Unlike previous approaches~\cite{jiang2022gatrust,ghafari2019dcat,li2025context} that manually define contextual factors (e.g., decomposing neighbor features into normal and abnormal components), our method incorporates contextual information through relational structures via context-aware meta-paths. This design offers three advantages: (i) It automatically captures latent contextual semantics from user-item interactions. (ii) It leverages meta-paths to connect different types of nodes and characterize diverse relations. (iii) It generalizes across networks by eliminating manual context engineering (as discussed in Section~\ref{use_case}).

\textbf{Time Embedding.}
Metapath2vec has a key limitation: it does not retain the temporal order of interactions within a meta-path. To address this problem and model trust dynamics, we introduce a time encoding function $\varphi: \Delta t \to \mathbb{R}^{d_t}$. Herein, $\Delta t$ represents the time gap between an interaction and its associated target node, and $d_t$ is the embedding dimension. We use relative time gaps instead of absolute timestamps to reduce inconsistencies that arise from the varying ranges of absolute timestamps (e.g., Unix timestamps) across different datasets and between training and testing sets. Additionally, unlike previous approaches~\cite{Xu2020Inductive,cong2022we} that focus on the target timestamp of each node, we assign each node’s timestamp as the time at which its earliest interaction occurs.
The rationale behind this choice is twofold: (i) The first interaction marks a node's entry into a network, which holds historical significance. (ii) It simplifies time encoding by using a single representative timestamp per node, thereby avoiding redundant computation. The purpose of $\varphi$ is to encode continuous time gaps into $d_t$-dimensional embeddings $h_t$, ensuring that temporally close gaps are also close in the embedding space. Formally, the time encoding function is expressed as:
\begin{equation}
    \varphi(\Delta t) = \cos{(\Delta t \cdot \omega + \delta)},
\end{equation}
where $\omega$ and $\delta$ are learnable parameters. The cosine function is chosen due to (i) its inherent periodicity, making it ideal to capture periodic temporal patterns, and (ii) its continuity and smoothness, which ensure that small changes in time result in small changes in the output of the function.

\textbf{Edge Attribute Embedding.}
Except for timestamps, each edge may have such attributes as trust levels and user-item ratings, which can improve prediction accuracy but are often ignored by existing dynamic heterogeneous GNNs. To address this, we encode these attributes into embedding vectors. For instance, if the attributes refer to trust levels $w$ and $w \in \{Trust,Distrust\}$, we first model them as $[0,1]^\top$ and $[1,0]^\top$ using one-hot encoding. Given that different datasets have different levels of trust, we then transform these one-hot vectors into embeddings $h_e \in \mathbb{R}^{d_e}$ of a specified dimension via a learnable matrix.

\subsection{Heterogeneous Attention Layer}
This layer aims to learn node embeddings by selectively propagating and aggregating trust information that encodes trust strength, relevance, and directionality. Specifically, we design a dual attention mechanism that assigns varying weights to node types (type attention) and nodes within each type (node attention), enabling CAT to discriminate the significance of heterogeneous interactions. Compared to existing attention mechanisms~\cite{han,hgt} for handling graph heterogeneity, the proposed mechanism offers two advantages: (i) It accounts for both temporal information and edge attributes, which are essential for accurate trust prediction but often overlooked in prior work. (ii) It does not rely on any meta-path and focuses on node type importance, reducing manual effort and improving generality. Furthermore, to enhance scalability, we adopt recent-time neighbor sampling and one-hop trust propagation strategies for controlling the amount of trust information and the range of its propagation. In this subsection, we denote $v_j$ as the source node and $v_i$ the target node.

To begin with, we define a message $m_{i,j}$ as the trust information that needs to be propagated from $v_j$ to $v_i$:
\begin{equation}
    \label{msg}
    m_{i,j} = h_j + h_t + h_e,
\end{equation}
where $h_j$, $h_t$, and $h_e$ represent the embeddings of $v_j$, relative time gaps, and edge attributes, respectively. The messages associated with $v_i$ serve as the basis for mining its latent features, such as the objective trust given by others. Note that we use an additive method rather than concatenation for message construction, as this avoids the high dimensionality introduced by concatenation and is thus more efficient.

\textbf{Type Attention.}
Given a specific node $v_i$, it can engage in various interactions with different types of nodes, each having distinct impacts. For example, for a user node, a neighboring user node from a $\langle user, user \rangle$ interaction propagates information about trust relationships, while a neighboring item node from a $\langle item, user \rangle$ interaction conveys information such as user preferences. Therefore, we employ a type attention mechanism to discriminate the significance of different node types or relations. First, we calculate the embedding for type $\tau$ by aggregating messages from nodes of type $\tau$ as:
\begin{equation}
    \label{h_tau}
    h_\tau = \sum_{j \in \mathcal{N}^{\tau}_{v_i}} \frac{1}{\sqrt{d_i d_j}} m_{i,j},
\end{equation}
where $\mathcal{N}^{\tau}_{v_i}$ denotes the set of neighboring nodes of type $\tau$ for $v_i$. $d_i$ and $d_j$ are the degrees of $v_i$ and $v_j$, used to normalize message influence based on node connectivity.

Then, we calculate the type attention score based on the node type embedding $h_\tau$ and the target node's embedding $h_{i}$. To obtain the relative importance of each type, we further normalize the attention scores across all the types using the softmax function. The above process is shown in Eq.~\ref{alpha_tau}.
\begin{equation}
    \label{alpha_tau}
    \begin{split}
    a_\tau = \sigma(\gamma^{\top}_{\tau} \cdot [h_i || h_\tau]),\\
    \alpha_\tau = \frac{\exp(a_\tau)}{\sum_{\tau^{\prime} \in \mathcal{T}} \exp(a_{\tau^{\prime}})},
    \end{split}
\end{equation}
where $\gamma_{\tau}$ is the attention vector specific to type $\tau$, $||$ is a concatenation operation, and $\sigma(\cdot)$ is a non-linear activation function such as LeakyReLU.

\textbf{Node Attention.} Falling into the same type, different nodes can influence a target node differently. Therefore, we further calculate node attention scores with the consideration of type importance. Specifically, given a node $v_i$ with type $\tau$ and its neighboring node $v_j$ with type $\tau^{\prime}$ (where $\tau$ and $\tau^{\prime}$ may be the same or different), the normalized node attention score is calculated using Eq.~\ref{beta_ij}.
\begin{equation}
    \label{beta_ij}
    \begin{split}
    b_{i,j} = \sigma(\eta^\top \cdot \alpha_{\tau^{\prime}} [h_i || m_{i,j}]),\\
    \beta_{i,j} = \frac{\exp(b_{i,j})}{\sum_{j^{\prime} \in \mathcal{N}_{v_i}} \exp(b_{i,j^{\prime}})},
    \end{split}
\end{equation}
where $\eta$ is the attention vector for nodes, and $\alpha_{\tau^{\prime}}$ refers to the significance of type $\tau^{\prime}$. The node attention score $\beta_{i,j}$ indicates the importance of $v_j$ to $v_i$. Note that the effects of time and edge attributes are also learned through the dual attention mechanism since they are encoded in the messages.

Finally, to derive the embedding of node $v_i$, we aggregate its related messages $m_{i,j}$, each weighted by $\beta_{i,j}$ (computed via sequential type and node attention mechanisms). After that, a fully-connected layer is applied to extract high-level features. The formal process is described below:
\begin{equation}
    \label{h_i}
    h_i = \sigma( (\sum_{j \in \mathcal{N}_{v_i}} \beta_{i,j} m_{i,j}) \cdot W  + B),
\end{equation} 
where $W$ and $B$ are learnable parameters.

\textbf{Neighbor Sampling.} A real-world heterogeneous network is typically very large and contains rich information. While this information benefits embedding learning, it may also introduce noise that undermines trust propagation and aggregation. Considering this issue and CAT's scalability, we adopt a neighbor sampling strategy~\cite{tgn,graphsage} that selects a subset of neighbors (or interactions) for trust propagation and aggregation. There are two sampling strategies: uniform sampling, which uniformly samples neighbors of a target node, and recent-time sampling, which selects neighbors having the newest interactions with the target node. We adopt the recent-time sampling strategy since it is more efficient, and new interactions are generally more important than historical ones~\cite{lin2021medley}. To the best of our knowledge, this strategy is applied for the first time in GNN-based dynamic trust prediction. A comparison of the two strategies is provided in Appendix~\ref{appendix_sampling}.

\textbf{One-Hop Trust Propagation.} Through the above process, latent trust information embedded in messages can be propagated and aggregated, guided by the graph structure. Specifically, if CAT is equipped with $L$ heterogeneous attention layers, $h_i$ fuses the information about trust from the $L$-hop neighbors of $v_i$. In other words, we can control the range of trust propagation by setting $L$. However, increasing $L$ introduces several issues. First, a large $L$ often leads to over-smoothing~\cite{chen2020measuring}, where node embeddings become indistinguishable. Second, deeper architectures significantly increase computational overhead, making them less scalable in large-scale trust networks. Moreover, previous work has shown that trust is conditionally transitive and tends to decay with distance~\cite{huo2023trustgnn,josang2007survey}, highlighting the importance of one-hop neighbors over multi-hop ones. Empirical results (Section~\ref{hyparameter_analysis}) further support the above statements: CAT performs best when $L=1$, indicating that trust information from one-hop neighbors is sufficient for accurate and efficient prediction. Therefore, we adopt a one-hop trust propagation strategy to enhance scalability without compromising accuracy.

\subsection{Prediction Layer} \label{prediction_layer}
This layer aims to predict the latent trust between any two user nodes within a specific context. Since the previous layer already generates expressive embeddings through the dual attention mechanism, this layer adopts a simple two-layer MLP as the prediction model. The input to the MLP is a combination of three types of embeddings: those of the trustor, the trustee, and the context. While embeddings of the trustor and trustee are directly obtained from the previous layer, the context embedding is not readily accessible. To enable context-aware trust prediction, we calculate a context embedding by averaging the embeddings of items within the same context. The reason is that items within the same context collectively characterize that context from multiple aspects, and averaging effectively preserves the general characteristics of the context. Alternative methods for generating context embeddings are discussed in Appendix~\ref{context_generator}. 
Formally, the prediction process is as follows:
\begin{equation}
    \label{w_ijk}
    \tilde{w}_{i,j,k} = \text{softmax}(\text{MLP}(h_i || h_j || h_k)),
\end{equation}
where $h_i$, $h_j$, and $h_k$ denote the embeddings of $v_i$, $v_j$, and the context $c_k$, respectively. $\tilde{w}_{i,j,k}$ is the probabilistic prediction vector of the trust from $v_i$ to $v_j$ within $c_k$. The predicted trust level is thus computed by $\bar{w}_{i,j,k} = \arg\max_q (\tilde{w}_{i,j,k})$, where $q$ refers to the index of the maximum value in $\tilde{w}_{i,j,k}$. It is important to note that $\bar{w}_{i,j,k}$ may not equal $\bar{w}_{j,i,k}$ due to the asymmetric nature of trust.

\textbf{Context-Aware Aggregation.} By the above steps, CAT has the potential to predict trust between users within any context, which is more fine-grained than the overall trust used in prior approaches. However, we are facing the lack of ground truth for trust levels within specific contexts. To address this challenge, we establish a connection between context-aware trust and overall trust. A solution is to average or sum trust levels across all contexts to derive an overall trust level~\cite{cdeeptrust}. This solution is efficient but does not consider the importance of different contexts. We thus introduce a context-aware aggregator to automatically determine the significance of each context in forming the overall trust level:
\begin{equation}
    \label{w_ij}
    \tilde{w}_{i,j} = \sum_{k=1}^{\left | \mathcal{C} \right |} g_k \cdot \tilde{w}_{i,j,k},
\end{equation}
where $\left | \mathcal{C} \right |$ refers to the number of contexts. $G_k = [g_1,\cdots,g_k,\cdots,g_{|\mathcal{C}|}]$ is a learnable vector, in which each element $g_k$ indicates the importance of a corresponding context, and the sum of the elements equals one via the softmax function. The output $\tilde{w}_{i,j}$ is the probabilistic prediction vector of the overall trust from $v_i$ to $v_j$. Similarly, we can compute the overall trust level using $\bar{w}_{i,j} = \arg\max_q (\tilde{w}_{i,j})$. The procedure of CAT is presented in Algorithm~\ref{algorithm}.

To summarize, \textit{CAT is the first GNN-based trust prediction model that not only predicts the overall trust but also provides insights into the contexts in which trust relationships are more likely to form, despite the absence of context-specific trust labels.} This is a key contribution that distinguishes CAT from previous approaches~\cite{jiang2022gatrust,ghafari2019dcat,li2025context}, which consider multiple contextual factors to guide embedding learning but cannot explicitly predict context-aware trust. This capability is significant for enabling flexible trust-based applications, such as fraud detection and access control.

\begin{algorithm}[!t]
\footnotesize
\SetCommentSty{small}
\LinesNumbered
\caption{CAT Procedure}
\label{algorithm}

\KwIn{A dynamic graph $\mathcal{G}(T)$;
}
\KwOut{Context-aware trust level $\bar{w}_{i,j,k}$, overall trust level $\bar{w}_{i,j}$;}

\Comment{\textbf{Embedding Initialization}}

Generate node embeddings $h_i, h_j$ for all nodes, time embeddings $h_t$ and edge attribute embeddings $h_e$ for all interactions;

\Comment{\textbf{Embedding Learning}}

\For{$v_i \in \mathcal{V}(T)$}
    {
    \For{$j \in \textnormal{Sampling} (\mathcal{N}_{v_i})$}
        {
        Construct a message $m_{i,j}$ using Eq.~\ref{msg};

        Calculate the type embedding $h_\tau$ using Eq.~\ref{h_tau};

        Calculate the type attention score $\alpha_\tau$ using Eq.~\ref{alpha_tau};

        Calculate the node attention score $\beta_{i,j}$ using Eq.~\ref{beta_ij};
        }
    Form $v_i$'s embedding $h_i$ using Eq.~\ref{h_i};
    }

\Comment{\textbf{Trust Relationship Prediction}}

\For{$e_{i,j} \in \mathcal{E}(T)$}
    {
    Calculate the context-aware trust vector $\tilde{w}_{i,j,k}$ using Eq.~\ref{w_ijk};

    Calculate the overall trust vector $\tilde{w}_{i,j}$ by aggregating trust across all contexts through Eq.~\ref{w_ij};

    Obtain the context-aware trust level $\bar{w}_{i,j,k}$ via $\arg\max_q (\tilde{w}_{i,j,k})$;

    Obtain the overall trust level $\bar{w}_{i,j}$ via $\arg\max_q (\tilde{w}_{i,j})$;
    }

\end{algorithm}

\begin{table*}[htbp]
\footnotesize
\centering
\caption{Statistics of Epinions, Ciao, and CiaoDVD datasets.}
\label{dataset}
\begin{tabular}{c|c|c|c|c|c|c|c}
\toprule[1.5pt]
Datasets       & \# Users & \# Items & \# Ratings & \# Trust relationships & \# Contexts & Timestamps of ratings & Timestamps of trust relationships       \\ \midrule
Epinions   & 9163    & 12573   & 265189    & 311158   & 25    & \ding{51}  & \ding{51} \\
Ciao   & 2378    & 16861   & 36065    & 57544   & 6    & \ding{51}  & \ding{55} \\
CiaoDVD   & 19533    & 16121   & 72665    & 40133   & 17    & \ding{51}  & \ding{55} \\
\bottomrule[1.5pt]
\end{tabular}
\vspace{-2mm}
\end{table*}

\subsection{Model Training}
To train CAT, we adopt the cross-entropy loss function to measure the discrepancy between the predicted overall trust and the ground truth. The loss function is formulated as:
\begin{equation}
    \label{loss_func}
    \mathcal{L} = - \sum_{e_{i,j} \in \mathcal{E}(T)} \log_{}{\tilde{w}_{i,j; w_{i,j}}} + \lambda \cdot \Vert \Theta \Vert^2_2,
\end{equation}
where $w_{i,j}$ denotes the ground truth, $\Theta$ denotes all learnable parameters of the model, and $\lambda$ controls the $L_2$ regularization to mitigate overfitting. Additionally, we use the Adam optimizer~\cite{adam} for updating the model parameters.
\section{Experimental Evaluation}
In this section, we first introduce experimental settings and then answer the following research questions: 
\textbf{RQ1:} How does CAT perform compared with the representative models regarding overall trust and context-aware trust?
\textbf{RQ2:} How does each component of CAT contribute to its performance?
\textbf{RQ3:} How do different hyperparameters affect the performance of CAT?
\textbf{RQ4:} Is CAT efficient and scalable compared with the representative models?
\textbf{RQ5:} Is CAT robust against data poisoning attacks compared with the representative models?
We also visualize the learned attention scores and embeddings to interpret CAT's behavior (Appendix~\ref{visualization_dual} and~\ref{appendix_emb}).

\subsection{Experimental Settings} \label{experimental_settings}
\textbf{Datasets.} 
Our experiments are conducted on three real-world datasets: Epinions~\cite{tang2015trust}, Ciao~\cite{tang2015trust}, and CiaoDVD~\cite{guo2014etaf}, all sourced from social networking-based consumer review sites. These datasets are widely used benchmarks for trust prediction and, to the best of our knowledge, are the only publicly available ones that provide temporal information, heterogeneous interactions, and item categories, making them well-suited for our study. Specifically, they include: (i) user-item ratings ranging from 1 to 5; (ii) overall trust relationships between users, e.g., user A trusts user B regardless of specific contexts; (iii) item categories, a type of contextual information; and (iv) interaction timestamps. The statistics of the datasets are given in Table~\ref{dataset}.

\textit{Data preparation.} 
For Epinions, we chronologically split the trust relationships into 70\%-15\%-15\% and 80\%-10\%-10\% for training, validation, and testing, two common split ratios used in dynamic graph analysis. For Ciao and CiaoDVD, where the trust relationships lack timestamps, we randomly split them into training, validation, and testing sets with ratios of $x$, $\frac{1-x}{2}$, and $\frac{1-x}{2}$, respectively ($x$ = 50\%, 60\%, 70\%, 80\%). In addition, we initialize the time embeddings of trust relationships as zero vectors to ensure CAT works correctly on these two datasets. Since only positive links (i.e., trusted relationships) exist in all datasets, an equal number of unlinked user pairs is randomly selected to form a distrusted set for training, validation, and testing. This method, known as negative sampling, is commonly used in current research to create a balanced dataset for effective model training~\cite{magic}.

\textbf{Baseline Models.} 
To conduct a comprehensive comparison, we select four types of GNN models as baselines that vary in their support for dynamicity and heterogeneity. (i) \textit{Homogeneous static model.} Guardian~\cite{lin2020guardian} is the first GNN model for trust prediction, which has demonstrated superiority over three types of non-GNN-based approaches. 
GATrust~\cite{jiang2022gatrust} utilizes attention mechanisms to incorporate multiple trust-related factors. (ii) \textit{Homogeneous dynamic model.} Medley~\cite{lin2021medley} and TrustGuard~\cite{trustguard} are two state-of-the-art GNN-based trust prediction models with dynamicity support. They employ continuous-time and discrete-time methods, respectively, to predict time-aware trust relationships. (iii) \textit{Heterogeneous static model.} HAN~\cite{han} introduces a hierarchical attention mechanism to learn the importance of meta-path-based neighbors and different meta-paths. KGTrust~\cite{KGTrust} is excluded from our evaluation due to its closed-source nature and limited improvements over Guardian.
(iv)~\textit{Heterogeneous dynamic model.} Research on GNN-based trust prediction has not yet evolved to this model type, while CAT belongs to it. To validate the effectiveness of CAT, we adapt HGT~\cite{hgt} to the trust prediction task. HGT is a continuous-time model that captures graph heterogeneity through meta-relations and learns mutual attention across them. 
In addition to these GNN-based models, we include Linear, a simple yet strong baseline that uses only linear layers but has been shown to outperform many complex GNNs~\cite{bilot2025sometimes}. For this baseline, node features are initialized using Metapath2vec with node-type information, given the absence of textual attributes in the datasets.

\textbf{Evaluation Metrics.} We evaluate the effectiveness of CAT and baselines using \textit{Mean Reciprocal Rank (MRR)}, \textit{Average Precision (AP)}, and \textit{Area Under the ROC Curve (AUC)}. These metrics are widely used to evaluate GNN-based link prediction~\cite{kipf2016variational,zhang2018link,li2024evaluating}, making them well-suited for our task. We do not use accuracy and F1 score, as they are more suitable for edge classification with explicit trust strength labels, whereas link prediction focuses on ranking positive links (trusted relationships) higher than negative ones (distrusted relationships). All metrics range from 0 to 1, with higher values indicating better performance. Detailed definitions are given in Appendix~\ref{metrics}. We also include \textit{running time} to assess model efficiency and scalability.

\textbf{Implementation Details.} 
We implemented CAT~\footnote{\url{https://github.com/Jieerbobo/CAT}} and baseline models using PyTorch on a server equipped with an Intel Xeon Platinum 8352V CPU and an RTX 4090 GPU. For training, we set the maximum number of epochs to 50 and adopt an early stopping strategy: training stops if AP on the validation set does not improve for 10 consecutive epochs. Regarding hyperparameters of CAT, we conducted a grid search and set the learning rate to 0.001, dropout to 0, $L_2$ regularization coefficient to 0.0005, embedding dimension to 64, batch size to 256, and trust propagation length to 1 by default. Baseline models were similarly tuned to achieve their best performance in trust prediction, except that their trust propagation length was fixed to 2 (a common choice in the literature~\cite{lin2021medley}). Considering the scales of the datasets, we set the number of sampled neighbors to 30 for Epinions and use all neighbors for Ciao and CiaoDVD. We report the average results obtained from 5 runs for each experiment.

\begin{table}[tbp]
	\centering 
	\footnotesize
	\caption{Performance comparison between CAT and different baselines on the Epinions dataset.}
        \label{rq1_epinions}
        {\vspace{-1mm}In each column, the best result is highlighted in \textbf{bold}, while the second-best result is \underline{underlined}.}
\\[2mm]
	\begin{tabular}{@{}c|ccc|ccc@{}}
		\toprule[1.5pt]
    \multirow{2.5}{*}{Models} &\multicolumn{3}{c|}{70\%-15\%-15\%} &\multicolumn{3}{c}{80\%-10\%-10\%}\\
    \cmidrule{2-7}
		&MRR &AP &AUC &MRR &AP &AUC \\
   
   \midrule
    \multicolumn{7}{c}{Task 1: Trust prediction for observed users}\\
   \midrule
             Linear &0.3866 &0.8247 & 0.8873 & 0.3622 &0.8305 &0.8907 \\
             Guardian &0.3097 &0.8237 & 0.9233 & 0.4979 &0.9080 &0.9444 \\
             GATrust &0.4380 &0.8851 & 0.9431 & 0.5168 &0.9135 &0.9460 \\
		 Medley &0.4762 &0.8944 &0.9440 &0.5577 &0.9336 &0.9628 \\
         TrustGuard &0.4955 &0.8919 &0.9382 &0.5390 &0.9214 &0.9542 \\
             HAN &0.4054 &0.8731 & 0.9396 & 0.4100 &0.8869 &0.9415 \\
		 HGT &\underline{0.5081} &\underline{0.9151} & \underline{0.9588} & \underline{0.6168} &\underline{0.9446} &\underline{0.9675} \\
		 \textbf{CAT} & \textbf{0.6025} &\textbf{0.9383} & \textbf{0.9677} &\textbf{0.6778} &\textbf{0.9603}&\textbf{0.9773} \\

   \midrule
    \multicolumn{7}{c}{Task 2: Trust prediction for unobserved users}\\
   \midrule
             Linear &0.2520 &0.9067 & 0.8047 & 0.2675 &0.9364 &0.8297 \\
             Guardian &0.1497 &0.8034 & 0.5902 & 0.1658 &0.8573 &0.6395 \\
             GATrust &0.1917 &0.8137 & 0.6017 & 0.1770 &0.8514 &0.6140 \\
		 Medley &0.1979 &0.8884 &0.7806 &0.2203 &0.9339 &0.8381 \\
         TrustGuard &0.2571 &0.8950 &0.7312 &0.2634 &0.9320 &0.7891 \\
             HAN &\underline{0.2707} &\underline{0.9365} & \underline{0.8773} & 0.2040 &0.9387 &0.8596 \\
		 HGT &0.2693 &0.9326 & 0.8603 & \underline{0.2931} &\underline{0.9504} &\underline{0.8633} \\
		 \textbf{CAT} & \textbf{0.4082} &\textbf{0.9527} & \textbf{0.8933} &\textbf{0.3987} &\textbf{0.9594}&\textbf{0.8799} \\

	\bottomrule[1.5pt]
	\end{tabular}
    \vspace{-3mm}
\end{table}

\begin{table*}[htbp]
	\centering 
	\footnotesize
	\caption{Performance comparison between CAT and different baselines on the Ciao and CiaoDVD datasets.}
    {\vspace{-1mm}In each column, the best result is highlighted in \textbf{bold}, while the second-best result is \underline{underlined}.}
\\[2mm]
        \label{rq1_ciao}
	\begin{tabular}{@{}c|c|ccc|ccc|ccc|ccc@{}}
		\toprule[1.5pt]
    \multirow{2.5}{*}{\makebox[0.05\textwidth][c]{Datasets}} &\multirow{2.5}{*}{Models} &\multicolumn{3}{c|}{50\%-25\%-25\%} &\multicolumn{3}{c|}{60\%-20\%-20\%} &\multicolumn{3}{c|}{70\%-15\%-15\%} &\multicolumn{3}{c}{80\%-10\%-10\%} \\
    \cmidrule{3-14}
		& &MRR &AP &AUC &MRR &AP &AUC &MRR &AP &AUC &MRR &AP &AUC \\
   
   \midrule
            \multirow{6}{*}{Ciao}
            &Linear &0.1978 &0.7676 & 0.7688 & 0.2118 &0.7836 &0.7847 &0.2026 &0.7730 & 0.7736 & 0.2098 &0.7817 &0.7824 \\
            &Guardian &0.2177 &0.8126 & 0.8318 & 0.2397 &0.8198 &0.8312 &0.2222 &0.8123 & 0.8290 & 0.2223 &0.8135 &0.8311 \\
            &GATrust &0.2475 &0.8292 & 0.8432 & 0.2440 &0.8285 &0.8440 &0.2438 &0.8313 & 0.8473 & 0.2293 &0.8236 &0.8433 \\
		&Medley &0.2092 &0.8049 &0.8257 &0.2356 &0.8270 &0.8438 &0.2573 &0.8406 & 0.8526 & 0.2484 &0.8358 &0.8513 \\
            &TrustGuard &0.2706 &0.8327 &0.8354 &0.2593 &0.8281 &0.8332 &0.2806 &0.8416 &0.8446 &0.2760 &0.8403 &0.8436 \\
            &HAN &0.2411 &0.8195 & 0.8299 & 0.2529 &0.8298 &0.8393 &0.2575 &0.8313 & 0.8408 & 0.2643 &0.8341 &0.8413 \\
		&HGT &\underline{0.2772} &\underline{0.8470} & \underline{0.8550} & \underline{0.2775} &\underline{0.8508} &\underline{0.8600} & \underline{0.2835} &\underline{0.8559} & \underline{0.8650} & \underline{0.3008} &\underline{0.8623} &\underline{0.8693} \\
		&\textbf{CAT} & \textbf{0.3716} &\textbf{0.9097} & \textbf{0.9215} &\textbf{0.3881} &\textbf{0.9149} &\textbf{0.9257} & \textbf{0.4150} &\textbf{0.9234} & \textbf{0.9327} &\textbf{0.4042} &\textbf{0.9200} &\textbf{0.9298} \\

  \midrule
            \multirow{6}{*}{CiaoDVD}
            &Linear &0.4119 &0.9168 & 0.9226 & 0.3995 &0.9114 &0.9176 &0.4097 &0.9156 & 0.9213 & 0.4157 &0.9174 &0.9231 \\
            &Guardian &0.3571 &0.9146 & 0.9358 & 0.3509 &0.9131 &0.9351 &0.3748 &0.9177 & 0.9366 & 0.3876 &0.9180 &0.9362 \\
            &GATrust &0.3848 &0.9199 & 0.9390 & 0.3796 &0.9187 &0.9379 &0.3891 &0.9193 & 0.9394 & 0.3660 &0.9126 &0.9331 \\
		&Medley &0.4318 &0.9320 &0.9440 &0.4605 &0.9394 &0.9492 &0.4578 &0.9392 & 0.9496 & 0.4742 &0.9410 &0.9497 \\
            &TrustGuard &0.5820 &0.9571 &0.9601 &0.5786 &0.9577 &0.9613 &0.5872 &0.9566 & 0.9586 & 0.5618 &0.9539 &0.9577 \\
            &HAN &0.7258 &0.9796 & 0.9816 & 0.7350 &0.9804 &0.9818 &0.7138 &0.9794 & 0.9819 & 0.7173 &0.9801 &0.9824 \\
		&HGT &\underline{0.7606} &\underline{0.9826} & \underline{0.9826} & \underline{0.7563} &\underline{0.9827} &\underline{0.9828} &\underline{0.7481} &\underline{0.9823} & \underline{0.9829} & \underline{0.7340} &\underline{0.9814} &\underline{0.9826} \\
		&\textbf{CAT} & \textbf{0.8225} &\textbf{0.9871} & \textbf{0.9861} &\textbf{0.8259} &\textbf{0.9872} &\textbf{0.9862} & \textbf{0.8413} &\textbf{0.9890} & \textbf{0.9883} &\textbf{0.8406} &\textbf{0.9890} &\textbf{0.9880} \\

	\bottomrule[1.5pt]
	\end{tabular}
    \vspace{-3mm}
\end{table*}

\subsection{Performance Comparison (RQ1)} \label{rq1}
In this subsection, we first evaluate the performance of CAT and baselines on predicting overall trust across three datasets, as these datasets only have labels for the overall trust. Subsequently, we illustrate how CAT outputs context-aware trust and discuss its practical significance. 

\textbf{Overall Trust.} 
For the Epinions dataset, where timestamps of trust relationships are available, we design two tasks to assess model performance on dynamic graphs: trust prediction for observed users (task~1) and unobserved users (task~2). A key characteristic of dynamic graphs is that users may exist (have interactions) or disappear (without interactions) as time evolves~\cite{grid}. To capture this, we define unobserved users as those who only exist in the validation and testing stages, with no interactions during training. As shown in Table~\ref{rq1_epinions}, CAT consistently performs better than baseline models across both tasks, regardless of data split ratios. The Linear baseline, though surpassing some homogeneous static models, lags far behind CAT and other baselines. These results highlight the necessity of incorporating heterogeneity and dynamicity into the model design. While HGT also considers these two aspects, it lacks support for trust properties and trust-related attributes, making it not fully compatible with trust prediction. Its complex architecture also risks overfitting, resulting in suboptimal performance. We also observe a clear performance decay for all models in task~2 compared with task~1, indicating that task~2 is more difficult. This is reasonable as task~2 requires that the patterns learned by the trust prediction models can be generalized to unobserved users. Despite the challenging nature of task~2, CAT improves MRR by 50.79\% and 36.03\% under the settings of 70\%-15\%-15\% and 80\%-10\%-10\%, respectively, compared with the best baseline. These findings showcase that CAT effectively handles dynamic graphs and has the potential to address the cold start issue~\cite{wang2020survey}, a long-standing issue in trust prediction. Additionally, we find that HAN performs worse than Medley and TrustGuard in task~1 but outperforms it in task~2. This suggests that modeling heterogeneity can deal with sparse trust relationships between unobserved users.

For the Ciao and CiaoDVD datasets, where timestamps of trust relationships are missing, we evaluate the stability of CAT and baselines by varying data split ratios. Table~\ref{rq1_ciao} shows that CAT consistently outperforms baselines across different split ratios, with average MRR improvements of 38.67\% on Ciao and 11.08\% on CiaoDVD over the best baseline. Among the baselines, HGT performs best, highlighting the significance of modeling both dynamicity and heterogeneity. Notably, heterogeneous GNNs show superiority over homogeneous GNNs, suggesting that heterogeneous graphs offer rich information for accurate trust prediction. Furthermore, the Linear baseline shows substantial performance gaps compared with most GNN models. This may be attributed to its overly simple architecture, which tends to underfit in dynamic heterogeneous networks. In contrast, CAT effectively captures diverse semantics through its advanced design, making it well-suited for trust prediction in such complex networks.

\begin{figure}[tbp]
    \centering
    \includegraphics[scale=0.33]{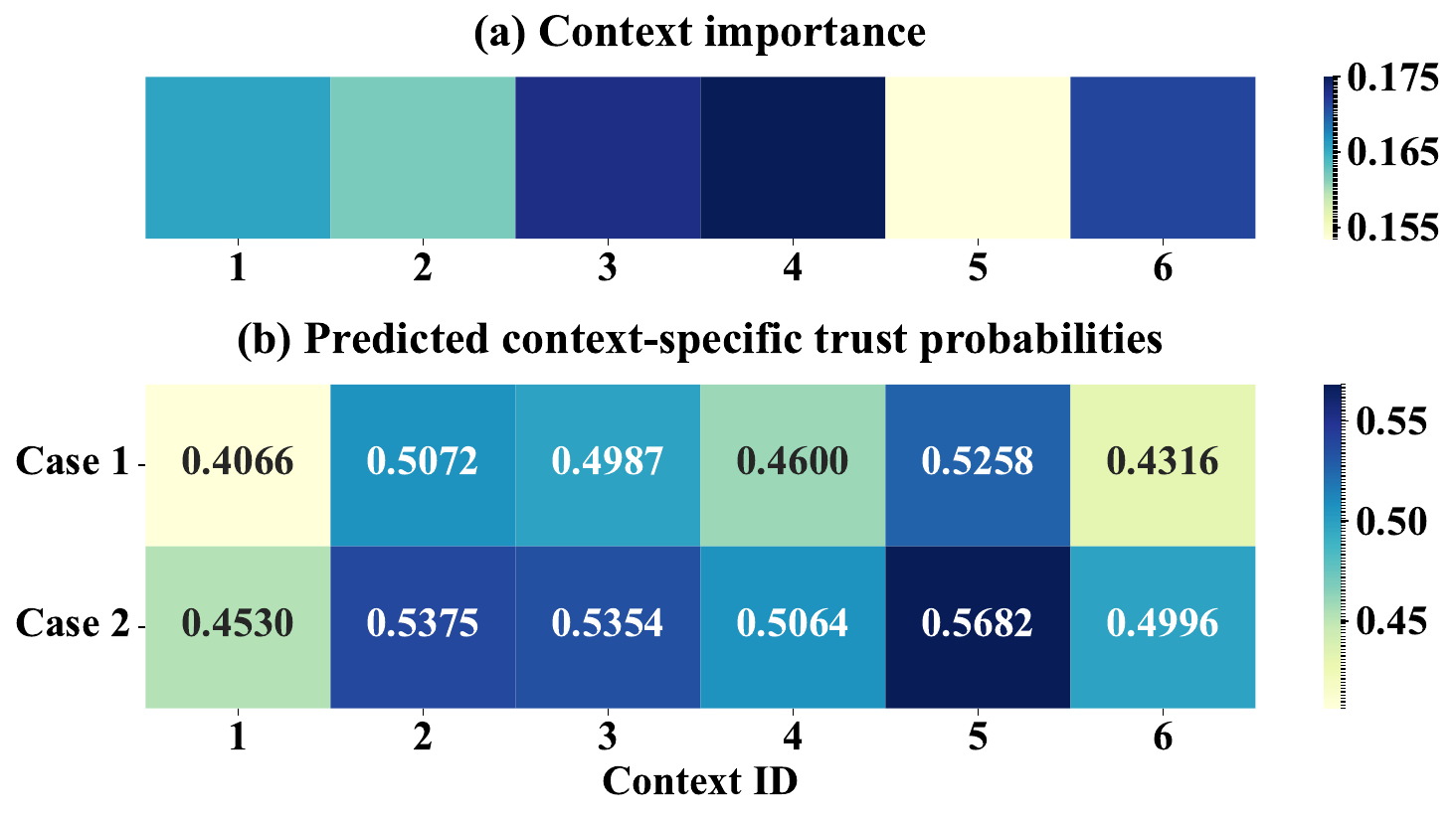}
    \vspace{-1mm}
    \caption{Results of context-aware trust prediction on the Ciao dataset. Case 1 illustrates a user pair $\langle v_i, v_j \rangle$ with an overall distrusted relationship, whereas case 2 depicts a user pair with an overall trusted relationship.}
    \label{rq1_context}
    \vspace{-3mm}
\end{figure}

\begin{table*}[tbp]
    \centering
    \footnotesize
    \begin{minipage}{0.70\textwidth}  
        \centering
        \caption{Ablation studies on CAT.}
        \label{rq2_ablation_study}
        \begin{tabular}{@{}c|ccc|ccc|ccc@{}}
		\toprule[1.5pt]
    \multirow{2.5}{*}{CAT and variants} &\multicolumn{3}{c|}{Epinions: Observed users} &\multicolumn{3}{c|}{Epinions: Unobserved users} &\multicolumn{3}{c}{Ciao} \\
    \cmidrule{2-10}
		&MRR &AP &AUC &MRR &AP &AUC &MRR &AP &AUC \\
   
   \midrule
            \textbf{CAT} &\textbf{0.6025} &\textbf{0.9383} & \textbf{0.9677} & \textbf{0.4082} &\textbf{0.9527} &\textbf{0.8933} & \textbf{0.4150} &\textbf{0.9234} &\textbf{0.9327} \\
            w/o Time Embedding &0.5575 &0.9240 & 0.9597 & 0.2441 &0.9040 &0.8045 & 0.3702 &0.9050 &0.9142 \\
		w/o Type Attention &0.5941 &0.9368 & 0.9677 & 0.3033 &0.9393 &0.8776 & 0.4035 &0.9216 &0.9322 \\
            w/o Node Attention &0.5781 &0.9328 & 0.9653 & 0.3975 &0.9501 &0.8851 & 0.4042 &0.9194 &0.9293 \\
            w/o Ca Meta-path &0.5763 &0.9274 &0.9600 &0.2924 &0.9328 &0.8592 & 0.3959 &0.9157 &0.9256 \\
            w/o Ca Aggregator &0.5642 &0.9285 & 0.9636 & 0.3859 &0.9480 &0.8817 & 0.3993 &0.9196 &0.9302 \\

	\bottomrule[1.5pt]
	\end{tabular}
    \end{minipage}%
    \hfill
    \begin{minipage}{0.25\textwidth}
        \vspace{2mm}
        \centering
        \footnotesize
        \captionsetup{aboveskip=2pt, belowskip=0pt}  
        \includegraphics[width=\textwidth]{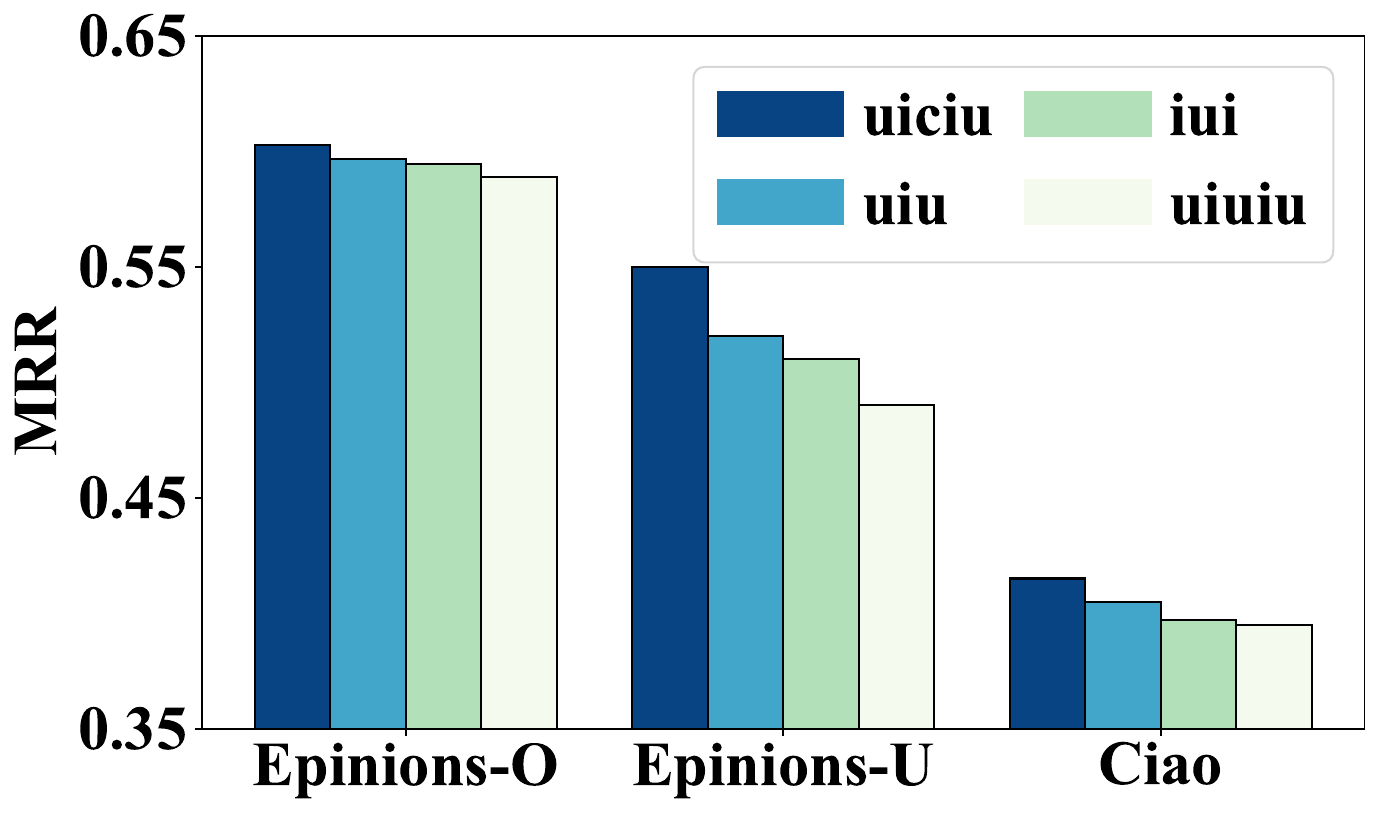}
        \captionof{figure}{Comparison of different meta-paths.}
        \label{metapath}
    \end{minipage}
    \vspace{-3mm}
\end{table*}

\textbf{Context-Aware Trust and a Real-World Use Case.}
Among existing GNN-based trust prediction models, CAT is unique in its ability to predict context-aware trust. Since other models lack this capability and current datasets do not provide context-specific trust labels, a direct quantitative comparison is not feasible. Instead, we present case studies to show how CAT outputs context-aware trust levels and their practical significance. 
First, CAT can explain how different contexts influence an overall trust level and provide a detailed trust probability for each context. Fig.~\ref{rq1_context}(a) shows the importance of each context learned by the context-aware aggregator, revealing that different contexts contribute differently to trust establishment. Fig.~\ref{rq1_context}(b) further illustrates that trust probabilities can vary notably for the same user pair across contexts, which aligns with the definition of context-awareness~\cite{sherchan2013survey}. For instance, in case 1, user $v_i$ generally distrusts $v_j$, yet exhibits trust in contexts 2 and 5 (trust probability $> 0.5$). Similarly, in case~2, $v_i$ generally trusts $v_j$ and has high trust probabilities in several contexts, but still distrusts $v_j$ in contexts 1 and 6. These examples highlight the importance of modeling context-awareness. However, baseline models that predict only overall trust fail to capture such nuanced contextual distinctions.

Second, CAT's context-awareness enables flexible trust-based applications, such as fraud detection on social platforms, where users may behave normally in some contexts but maliciously in others (e.g., posting benign reviews for electronics but fake reviews for clothing). CAT models the social network as a contextual trust graph and predicts trust for each user pair within each context. As shown in Fig.~\ref{rq1_context}(b), when CAT's predicted trust probability for $v_j$ falls below a threshold in certain contexts, this indicates potential fraudulent behavior in those contexts. Such signals allow the platform to warn $v_i$ or restrict $v_j$'s activities there. The threshold can be customized based on individual trust preferences, thus providing flexibility.

\subsection{Ablation Study (RQ2)} \label{ablation_study}
In this subsection, we investigate the necessity and rationality of each component in CAT using a fixed 70\%-15\%-15\% split ratio across Epinions and Ciao datasets, which have the largest and smallest sizes, respectively. 

To begin with, we construct five CAT variants by removing key components: 
(i) CAT w/o Time Embedding, which ignores temporal information and involves changes in both embedding and heterogeneous attention layers. 
(ii) CAT w/o Type Attention, which assumes equal importance across node types.
(iii) CAT w/o Node Attention, which assumes equal importance among nodes within the same type.
(iv) CAT w/o Ca Meta-path, where user and item embeddings are initialized with random vectors instead of using the context-aware meta-path.
(v) CAT w/o Ca Aggregator, which replaces the context-aware aggregator with a mean aggregator that averages trust across all contexts to produce overall trust.

It can be concluded from Table~\ref{rq2_ablation_study} that: (i) CAT is consistently better than its five variants, validating the importance of integrating trust dynamicity, type and node importance under graph heterogeneity, and context-awareness. This joint consideration is a unique contribution of our approach, making CAT effective for trust prediction and applicable to real-world networks.
(ii) Time embeddings and context-aware meta-paths contribute the most to model performance, indicating that basic trust properties are crucial for accurate trust prediction. 
(iii) Neither ``w/o Node Attention'' nor ``w/o Type Attention'' is competitive to CAT, highlighting the significance of node distinctions and semantic information in forming effective embeddings. 
(iv) Removing the context-aware aggregator reduces performance, confirming that different contexts influence trust differently. Effectively modeling such contextual differences can improve overall trust prediction.

We further justify the introduction of the new context-aware meta-path $uiciu$ in Fig.~\ref{metapath}. The results show that $uiciu$ has superiority over other meta-paths thanks to its incorporation of contextual information. Additionally, $uiuiu$, which captures high-hop trust information, is less effective than $uiu$, suggesting that irrelevant information from distant connections may undermine trust propagation and aggregation.

\begin{figure}[tbp]
    \centering
    \subfigcapskip=-3pt

    \subfigure[Epinions]{\includegraphics[scale=0.27]{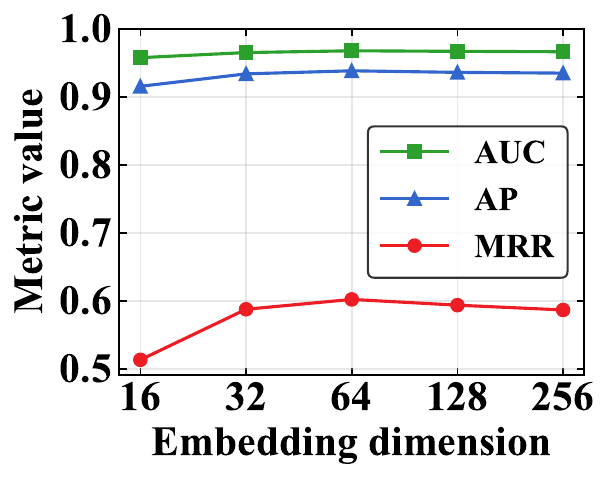}\label{rq3_emb1}}
    \subfigure[Epinions]{\includegraphics[scale=0.27]{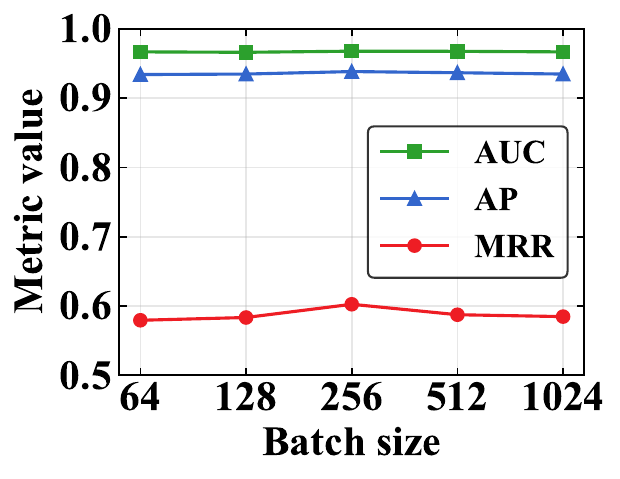}\label{rq3_batch1}}
    \subfigure[Observed users]{\includegraphics[scale=0.27]{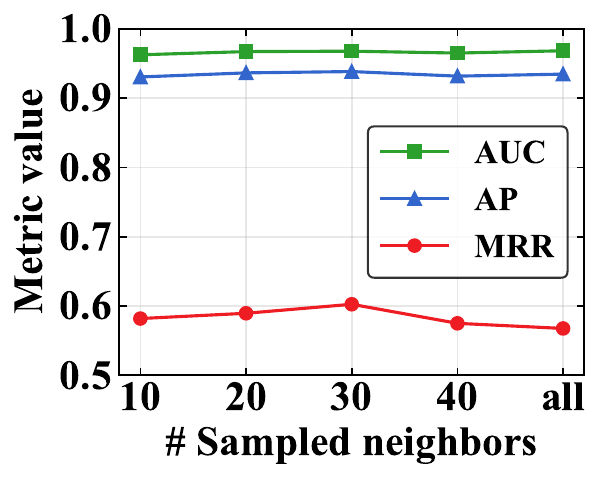}\label{rq3_sample1}}

    \vspace{-2mm}
    
    \subfigure[Ciao]{\includegraphics[scale=0.27]{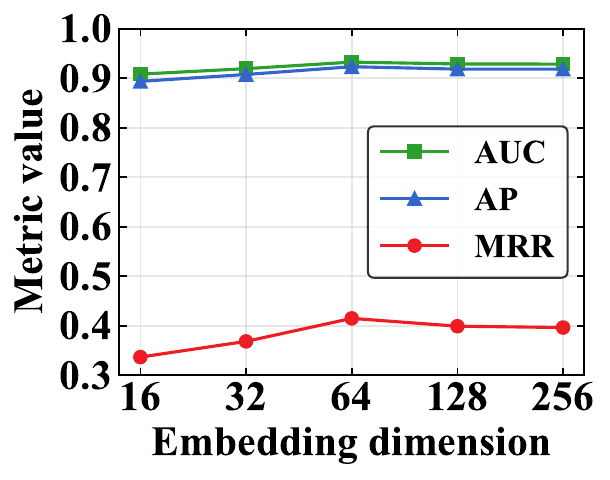}\label{rq3_emb2}}
    \subfigure[Ciao]{\includegraphics[scale=0.27]{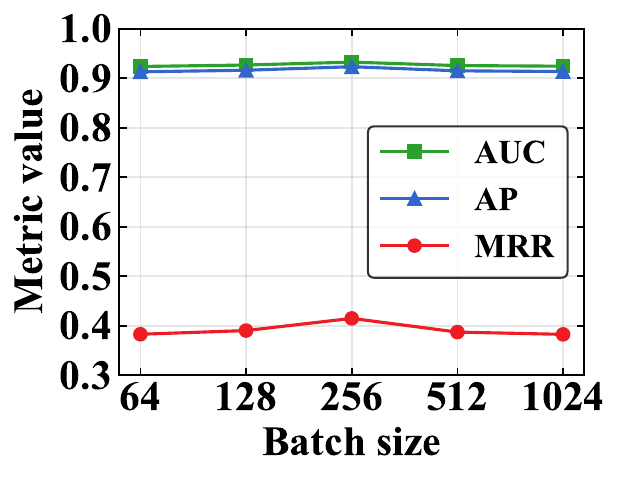}\label{rq3_batch2}}
    \subfigure[Unobserved users]{\includegraphics[scale=0.27]{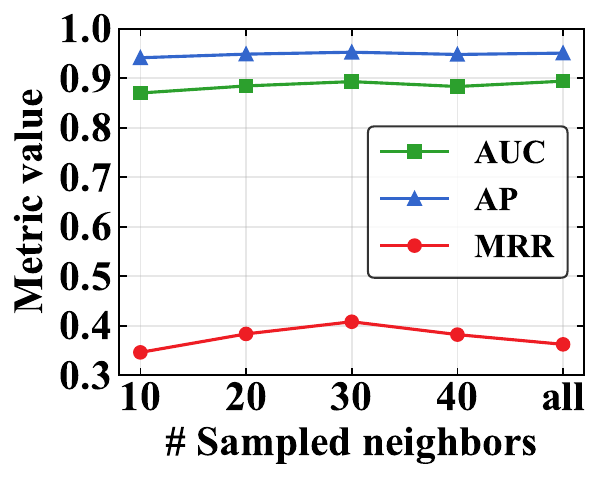}\label{rq3_sample2}}
    \vspace{-2mm}
    \caption{Effects of embedding dimension, batch size, and number of sampled neighbors on model performance.}
    \vspace{-2mm}
\end{figure}

\begin{figure}[tbp]
    \centering
    \subfigcapskip=-3pt
    \subfigure{\includegraphics[scale=0.35]{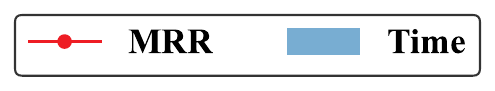}}
    \vspace{-4mm}

    \setcounter{subfigure}{0}
    \subfigure[Epinions]{\includegraphics[scale=0.36]{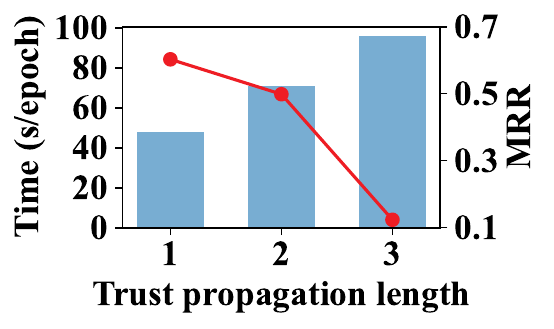}\label{rq3_length1}}
    \subfigure[Ciao]{\includegraphics[scale=0.36]{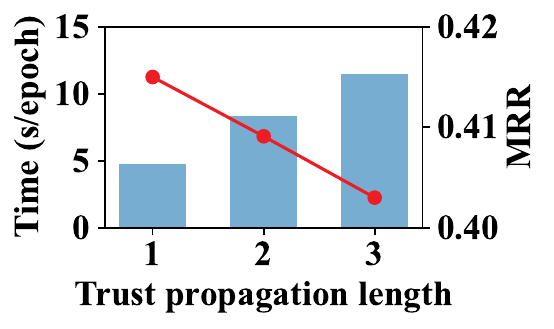}\label{rq3_length2}}
    \vspace{-2mm}
    \caption{Effect of trust propagation length on model performance.}
    \label{rq3_length}
    \vspace{-3mm}
\end{figure}

\begin{figure*}[tbp]
    \centering
    \subfigcapskip=-3pt

    \subfigure[Ciao]{\includegraphics[scale=0.28]{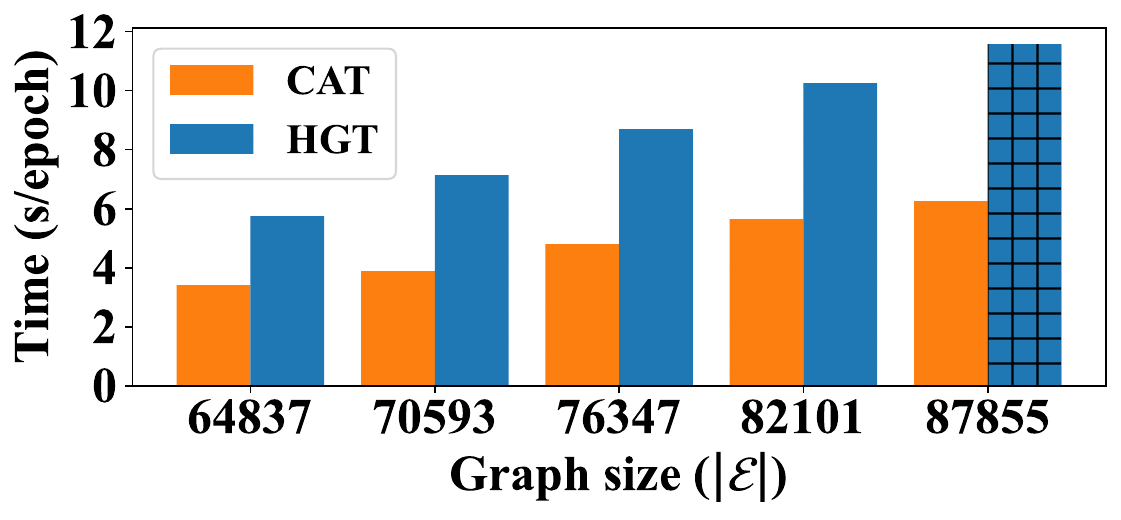}\label{rq4_scalability1}}\hspace{2mm}
    \subfigure[CiaoDVD]{\includegraphics[scale=0.28]{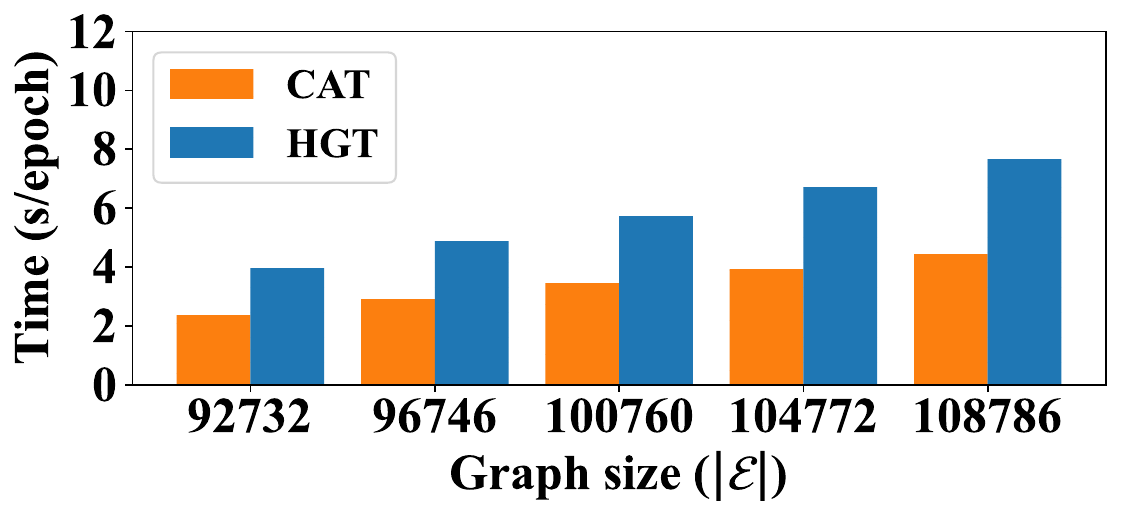}\label{rq4_scalability3}}\hspace{2mm}
    \subfigure[Epinions]{\includegraphics[scale=0.28]{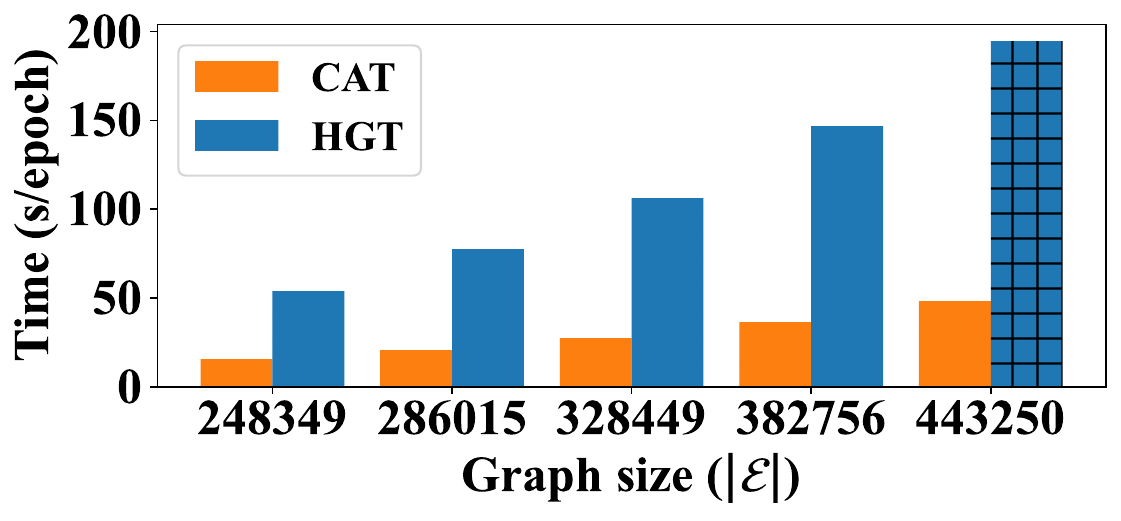}\label{rq4_scalability2}}
    \vspace{-2mm}
    \caption{Running time of CAT and HGT on various graph sizes.}
    \label{rq4_scalability}
    \vspace{-3mm}
\end{figure*}

\subsection{Hyperparameter Analysis (RQ3)} \label{hyparameter_analysis}
In this subsection, we analyze the impact of embedding dimension, batch size, number of sampled neighbors, and trust propagation length on CAT's performance using a fixed 70\%-15\%-15\% split ratio across Epinions and Ciao. 

\textbf{Embedding Dimension.} As shown in Fig.~\ref{rq3_emb1} and Fig.~\ref{rq3_emb2}, CAT's performance improves as the embedding dimension increases. However, excessively increasing the embedding dimension results in a performance decline. This outcome is as expected since a small embedding dimension limits the model's learning ability, while a large dimension increases the risk of overfitting. 

\textbf{Batch Size.} A real-world network is often extremely large, making it impractical to train a GNN-based trust prediction model by directly inputting the entire network. A solution is to divide the large network into subgraphs and process several subgraphs at a time. The batch size, referring to the number of subgraphs processed together, may impact the model's performance. A small batch size results in a partial view of node behaviors, leading to reduced prediction accuracy. Instead, a large batch size enhances the model performance by covering more information. Fig.~\ref{rq3_batch1} and Fig.~\ref{rq3_batch2} show that CAT is not sensitive to batch size. 

\textbf{Number of Sampled Neighbors.} The number of sampled neighbors determines subgraph size and affects scalability. We study its impact on model performance using the Epinions dataset, since the smaller Ciao and CiaoDVD datasets do not incur significant computational costs when all neighbors are used. As shown in Fig.~\ref{rq3_sample1} and Fig.~\ref{rq3_sample2}, model performance initially improves with more sampled neighbors but eventually decreases. This is reasonable because sampling too few neighbors leads to information loss, whereas sampling too many neighbors may introduce noise that degrades performance. 

\textbf{Trust Propagation Length.} Trust propagation length refers to the number of neighbor hops used to form a node's embedding. As we can see in Fig.~\ref{rq3_length1} and Fig.~\ref{rq3_length2}, CAT's performance significantly decreases while running time increases with longer propagation lengths, differing from prior findings~\cite{trustguard,huo2023trustgnn} where 2 or 3 hops are preferred. The reasons are two-fold: (i) From a data perspective, we focus on a heterogeneous graph while those studies use homogeneous ones. Although the heterogeneous graph provides rich semantics, it also introduces noise. Compared with one-hop neighbors, high-hop neighbors are less crucial and may mislead model training~\cite{cong2022we}. (ii) From a model perspective, the embeddings initialized by Metapath2vec already cover high-hop information. Thus, there is no need to stack multiple heterogeneous attention layers for trust propagation. Despite using one-hop neighbors, CAT still has high accuracy (see Section~\ref{rq1}).

\subsection{Scalability Analysis (RQ4)} \label{scalability_analysis}
In this subsection, we evaluate CAT's scalability by comparing its running time (i.e., average training time per epoch) with that of HGT, the best baseline that also considers dynamicity and heterogeneity, across varying graph sizes. A detailed time complexity analysis of CAT is provided in Appendix~\ref{time_complexity}.

Fig.~\ref{rq4_scalability} illustrates that as the graph size increases, the running time of both models increases for all datasets. Notably, CAT achieves average reductions in running time of 44.39\% and 40.79\% on the Ciao and CiaoDVD datasets, respectively, compared to HGT. Both models run faster on CiaoDVD than on Ciao because Ciao contains more trust relationships, which are the target links for prediction. On the Epinions dataset, CAT exhibits a more pronounced efficiency advantage, requiring about $48s$ per epoch for training on a graph with $443K$ edges, compared to $194s$ for HGT, yielding an average reduction of 73.97\% across different graph sizes. This high efficiency stems from CAT's one-hop trust propagation and recent-time neighbor sampling strategies. The former restricts trust propagation to immediate neighbors, reducing computation while maintaining high accuracy (see Fig.~\ref{rq3_length}). The latter focuses on limited yet crucial interactions. For comparison, considering interactions with all neighbors makes CAT nearly three times slower. Together, these strategies greatly reduce the number of messages to be propagated and improve scalability, making CAT suitable for practical deployment.

\begin{table*}[tbp]
\centering 
\footnotesize
\caption{Robustness comparison under trust-oriented and GNN-oriented attacks (MRR results).}
\label{robustness_all}
{\vspace{-1mm}\ding{172}/\ding{173}: trust predictions for observed/unobserved users. $p$: perturbation rate\\(ratio of added adversarial links to original links). MDR: Maximum Drop Rate (lower is better).}
\\[2mm]

\begin{tabular}{@{}c|c|c|ccccc|ccccc@{}}
\toprule[1.5pt]
\multirow{2.5}{*}{Tasks} & \multirow{2.5}{*}{Models} & \multirow{2.5}{*}{Clean} 
& \multicolumn{5}{c|}{Trust-oriented Attacks} 
& \multicolumn{5}{c}{GNN-oriented Attacks} \\
\cmidrule{4-8} \cmidrule{9-13}
& & & $p$=5\% & $p$=10\% & $p$=15\% & $p$=20\% & MDR $\downarrow$ & $p$=5\% & $p$=10\% & $p$=15\% & $p$=20\% & MDR $\downarrow$ \\
\midrule
\multirow{3}{*}{\ding{172}} 
& Medley        & 0.4762 & 0.4552 & 0.4367 & 0.4155 & 0.4079 & 14.34\% & 0.4650 & 0.4576 & 0.4477 & 0.4393 & 7.75\% \\
& TrustGuard    & 0.4955 & 0.4831 & 0.4820 & 0.4565 & 0.4528 & 8.62\%  & 0.4813 & 0.4711 & 0.4664 & 0.4721 & 5.87\% \\
& \textbf{CAT}  & 0.6025 & 0.5968 & 0.6046 & 0.6144 & 0.6070 & \textbf{0.95\%} & 0.5999 & 0.5869 & 0.5842 & 0.5821 & \textbf{3.39\%} \\
\midrule
\multirow{3}{*}{\ding{173}} 
& Medley        & 0.1979 & 0.1894 & 0.1791 & 0.1756 & 0.1715 & 13.34\% & 0.1909 & 0.1890 & 0.1846 & 0.1737 & 12.23\% \\
& TrustGuard    & 0.2571 & 0.2465 & 0.2431 & 0.2253 & 0.2250 & 12.49\% & 0.2477 & 0.2296 & 0.2348 & 0.2289 & 10.97\% \\
& \textbf{CAT}  & 0.4082 & 0.3988 & 0.3924 & 0.3911 & 0.3893 & \textbf{4.63\%} & 0.4022 & 0.4029 & 0.3918 & 0.3858 & \textbf{5.49\%} \\
\bottomrule[1.5pt]
\end{tabular}

\vspace{-3mm}
\end{table*}

\subsection{Robustness Analysis (RQ5)} \label{robustness_analysis}
In this subsection, we evaluate the robustness of CAT and two state-of-the-art GNN-based trust prediction models (i.e., Medley~\cite{lin2021medley} and TrustGuard~\cite{trustguard}) against data poisoning attacks. Specifically, we follow TrustGuard~\cite{trustguard} to construct trust-oriented attacks and T-Spear~\cite{lee2024spear}, the only recent data poisoning attack targeting continuous-time GNNs, to perform GNN-oriented attacks. Experiments are conducted on the Epinions dataset, which contains timestamped trust relationships and is thus compatible with T-Spear. Due to space constraints, we report only MRR results. Please refer to Appendix~\ref{attack_setting} for comprehensive results and detailed attack settings.

As shown in Table~\ref{robustness_all}, all models exhibit performance declines as the perturbation rate increases, but CAT outperforms the baselines under both attacks. For example, under trust-oriented attacks, CAT experiences a maximum performance drop of 0.95\% in task~\ding{172}, compared to 14.34\% for Medley and 8.62\% for TrustGuard. Similarly, in task~\ding{173}, CAT's maximum performance drop is 4.63\%, smaller than Medley's 13.34\% and TrustGuard's 12.49\%.
We observe that CAT is more robust against trust-oriented attacks than GNN-oriented ones, while Medley and TrustGuard show the opposite trend. This may be because the surrogate model used in T-Spear is closer to CAT's architecture than those of the other two models. Additionally, we note an interesting phenomenon: under task~\ding{172} of trust-oriented attacks, CAT's performance initially declines but later improves, even surpassing the clean setting. We analyze this result at a 15\% perturbation rate and find that this attack may act as data augmentation for two reasons: (i) It substantially reduces the number of connected components (from 100 to 2) with a small change in the clustering coefficient (0.133 vs. 0.131). This enhances graph connectivity and thus facilitates trust propagation. (ii) Similarity distribution analysis shows a 25.68\% overlap between adversarial and original links, implying that some adversarial links are structurally benign.

\begin{figure}[tbp]
    \centering
    \includegraphics[scale=0.3]{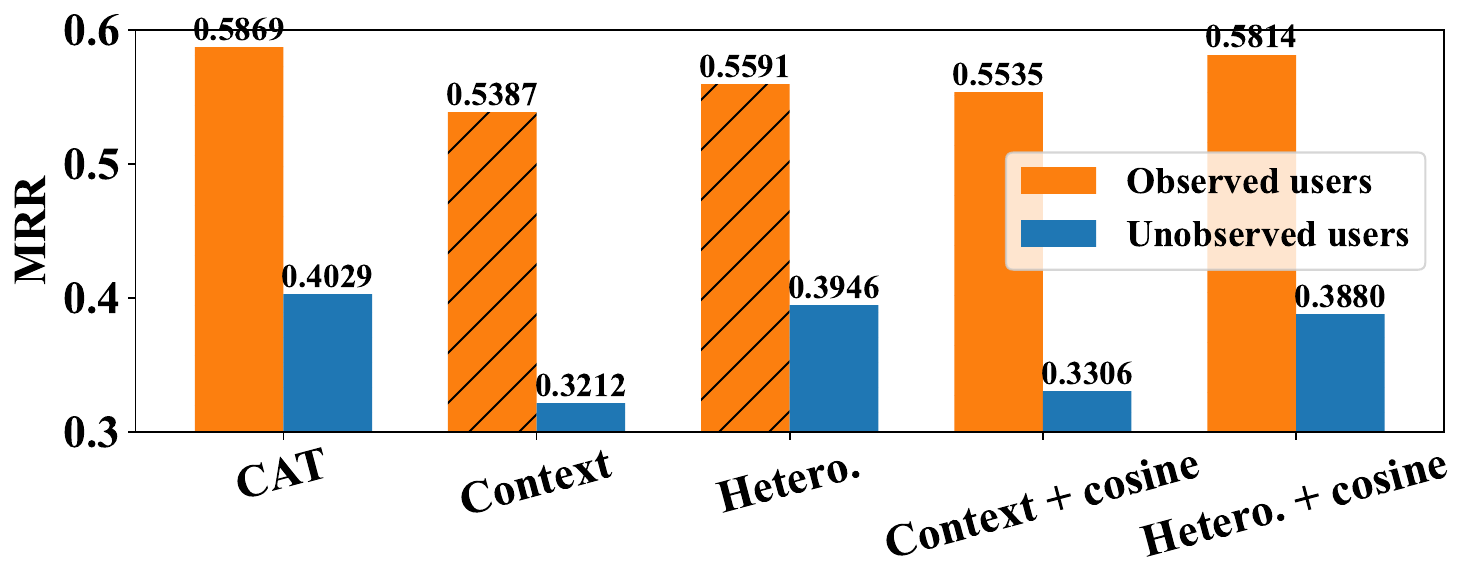}
    \vspace{-1mm}
    \caption{Detailed evaluation and comparison of robustness against GNN-oriented attacks with 10\% perturbation.}
    \label{robustness_poison}
    \vspace{-3mm}
\end{figure}

To summarize, CAT demonstrates superior robustness over existing GNN-based trust prediction models, owing to its strong semantic understanding. By learning from a heterogeneous graph and incorporating advanced designs, CAT gains a comprehensive view of intrinsic interactions and node semantics, thereby increasing the difficulty for attackers to succeed. For instance, T-Spear's surrogate model overlooks such semantic diversity. To validate this viewpoint, we construct two variants: (i) ``Context'', which disables heterogeneity modeling, and (ii) ``Hetero.'', which weakens contextual understanding. We also develop a defense mechanism for comparison because of the limited work in this area. Inspired by previous work~\cite{trustguard,zhang2020gnnguard}, we employ cosine similarity to filter out suspicious links. Fig.~\ref{robustness_poison} shows the MRR results under GNN-oriented attacks at the perturbation rate of 10\%. The results indicate that both heterogeneity and context-awareness are critical to CAT's robustness, with the ability to handle heterogeneity being particularly significant. While cosine similarity offers slight gains, it remains inferior to CAT.
\section{Discussion} \label{discussion}
\textbf{Generality to Different Networks.} \label{use_case}
While we only demonstrated CAT's effectiveness in social networks, it is also applicable to other real-world networks that can be abstracted into \textit{user}, \textit{item}, and \textit{item category} (i.e., \textit{context}). Examples include financial and employment networks. The reasons are as follows: 
(i) Trust issues are prevalent in these networks, making trust prediction critical. For instance, in financial networks, fraudulent activities are a serious concern. CAT can analyze complex node interactions to uncover latent trust relationships and mitigate security threats.
(ii) These networks are highly dynamic, with interactions evolving over time. CAT is effective at capturing such dynamics to predict time-aware trust relationships.
(iii) These networks are heterogeneous in nature, with diverse node and edge types. In financial networks, \textit{user} can be investors or financial advisors, \textit{item} can be stocks or cryptocurrencies, and \textit{context} denotes the type of financial product. CAT can effectively handle such heterogeneity and predict the trust an investor has in an advisor for recommending a specific product type.
Furthermore, the proposed context-aware meta-path remains applicable to these networks, reducing the need for additional human effort.

\textbf{Other Types of Adversarial Attacks.}
We evaluated CAT's robustness against two data poisoning attacks in Section~\ref{robustness_analysis}. Here, we further discuss its vulnerabilities to adaptive attacks and evasion attacks. 
First, consider an adaptive attacker who is aware that CAT derives its robustness from contextual differentiation and aims to blur contextual boundaries, making the model misinterpret distinct contexts as similar. Specifically, this adaptive data poisoning attack allows colluding attackers to manipulate interactions across contexts, e.g., by injecting malicious user-item interactions with timestamps crafted to mimic normal behavior. We evaluate CAT against this strong attack in Appendix~\ref{appendix_adaptive}. The results show that while this attack is more disruptive than trust-oriented and GNN-oriented attacks, CAT remains robust, particularly when predicting trust for observed users. This robustness stems from its context-aware aggregation mechanism, which adaptively weights contexts to mitigate noise and preserve contextual distinctions.

Second, evasion attacks can cause a model to misclassify a pair's trust relationship. Edge filtering~\cite{wu2019adversarial} based on Jaccard or cosine similarities, as well as low-rank based defenses~\cite{entezari2020all,mu2021hard}, can be employed to filter out suspicious links before they are fed into CAT for prediction. Additionally, adversarial training~\cite{adversarial_training} improves model robustness by augmenting the training data with adversarial examples. In summary, these defenses can be easily integrated into CAT to further enhance its resilience against various adversarial attacks.

\textbf{Diverse Trust Prediction Tasks.}
Beyond predicting trust relationships between trustor-trustee pairs, which corresponds to edge classification or link prediction in the field of GNNs, CAT also supports node-level trust prediction (i.e., node classification) and community-level trust prediction (i.e., graph classification). For node-level trust prediction, CAT's prediction layer takes a node's embedding along with a context embedding as input and outputs the node's trust within the given context. For community-level trust prediction, it takes a group of nodes' embeddings along with a context embedding as input and outputs the group's trust within that context. This support for diverse tasks makes CAT flexible in practice.

\textbf{Limitations and Future Work.}
While CAT extracts rich semantics through Metapath2vec and a dual attention mechanism, it currently overlooks textual attributes, such as user reviews and profiles, which may also exist in real-world networks. These texts can reveal user preferences and characteristics that are closely related to trust establishment. We did not include this information primarily due to scalability concerns and the limited availability of high-quality textual data in existing datasets. Nevertheless, CAT's modular design allows easy integration of plug-in text encoders, including lightweight pre-trained models and LLM-based feature extractors, enabling future incorporation of textual information. For example, recent years have witnessed a surge in applying LLMs across various domains owing to their extensive knowledge bases and strong reasoning capabilities~\cite{liu2024exploring,heharnessing}. By prompting LLMs to extract crucial information from the texts, we can obtain informative embeddings. This approach offers two advantages: (i) It enables CAT to comprehensively model real-world networks, improving its practicality. (ii) LLM-enhanced node embeddings offer superior robustness against both poisoning and evasion attacks~\cite{guo2024learning,zhang2024can}, further enhancing CAT's robustness. 
\section{Conclusion}
In this paper, we proposed CAT, a GNN-based trust prediction model that captures trust dynamicity, represents real-world heterogeneity, supports context-awareness, and is robust against data poisoning attacks. To incorporate contextual and temporal information, we introduced a context-aware meta-path and utilized a time encoding function, respectively. A dual attention mechanism identifies the importance of diverse interactions, while recent-time neighbor sampling and one-hop trust propagation strategies enhance scalability. By generating context embeddings and linking context-aware trust with overall trust, CAT can predict both types of trust. Extensive experiments on three real-world datasets demonstrate CAT's effectiveness, scalability, and robustness.


\section*{Acknowledgment}
This work is supported in part by the National Natural Science Foundation of China under Grant U23A20300; in part by the Key Research Project of Shaanxi Natural Science Foundation under Grant 2023-JC-ZD-35; in part by the Concept Verification Funding of Hangzhou Institute of Technology of Xidian University under Grant GNYZ2024XX007; in part by the 111 Center under Grant B16037; and in part by the Fundamental Research Funds for the Central Universities under Grant YJSJ25011.

\bibliographystyle{IEEEtran}
\bibliography{ref}

\appendices

\section{Time Complexity of CAT} \label{time_complexity}
We analyze the time complexity of the heterogeneous attention layer, as it is the dominant component in CAT. Let $|\mathcal{V}|$ be the number of nodes, $|\mathcal{E}|$ the number of edges, $|\mathcal{T}|$ the number of node types, and $d$ the embedding dimension. At first, the time complexities for calculating type attention and node attention are $O(|\mathcal{T}|d + |\mathcal{E}|d)$ and $O(|\mathcal{E}|d)$, respectively. Then, the message aggregation for forming node embeddings, as shown in Eq.~\ref{h_i}, costs about $O(|\mathcal{V}|d^2 + |\mathcal{E}|d)$. Therefore, the overall time complexity of CAT is about $O(|\mathcal{V}|d^2+|\mathcal{E}|d+|\mathcal{T}|d)$. Since we adopt a neighbor sampling strategy, the actual used edges (i.e., interactions) would be smaller than $|\mathcal{E}|$. Additionally, the one-hop trust propagation strategy makes the complexity independent of the number of GNN layers $L$. Given that $|\mathcal{T}| \ll |\mathcal{E}|$, the total time complexity of CAT is correlated to $|\mathcal{V}|$, $|\mathcal{E}|$, and $d$.

\begin{table*}[tbp]
\centering 
\footnotesize
\caption{Robustness comparison under trust-oriented and GNN-oriented attacks (AP and AUC results).}
\label{robustness_apauc}
{\vspace{-1mm}\ding{172}/\ding{173}: trust predictions for observed/unobserved users. $p$: perturbation rate\\(ratio of added adversarial links to original links). MDR: Maximum Drop Rate (lower is better).}
\\[2mm]

\begin{tabular}{@{}c|c|c|c|ccccc|ccccc@{}}
\toprule[1.5pt]
\multirow{2.5}{*}{Metrics} & \multirow{2.5}{*}{Tasks} & \multirow{2.5}{*}{Models} & \multirow{2.5}{*}{Clean} 
& \multicolumn{5}{c|}{Trust-oriented Attacks} 
& \multicolumn{5}{c}{GNN-oriented Attacks} \\
\cmidrule{5-9} \cmidrule{10-14}
& & & & $p$=5\% & $p$=10\% & $p$=15\% & $p$=20\% & MDR $\downarrow$ & $p$=5\% & $p$=10\% & $p$=15\% & $p$=20\% & MDR $\downarrow$ \\
\midrule

\multirow{6}{*}{AP}
& \multirow{3}{*}{\ding{172}} 
& Medley        & 0.8944 & 0.8862 & 0.8855 & 0.8728 & 0.8613 & 3.70\% & 0.8923 & 0.8838 & 0.8827 & 0.8810 & 1.50\% \\
& & TrustGuard    & 0.8919 & 0.8860 & 0.8855 & 0.8724 & 0.8685 & 2.62\%  & 0.8853 & 0.8816 & 0.8826 & 0.8814 & 1.18\% \\
& & \textbf{CAT}  & 0.9383 & 0.9394 & 0.9391 & 0.9415 & 0.9401 & \textbf{-0.09\%} & 0.9399 & 0.9350 & 0.9326 & 0.9335 & \textbf{0.61\%} \\
\cmidrule{2-14}
& \multirow{3}{*}{\ding{173}} 
& Medley        & 0.8884 & 0.8714 & 0.8473 & 0.8430 & 0.8328 & 6.26\% & 0.8799 & 0.8758 & 0.8780 & 0.8652 & 2.61\% \\
& & TrustGuard    & 0.8950 & 0.8918 & 0.8918 & 0.8824 & 0.8802 & 1.65\% & 0.8901 & 0.8881 & 0.8898 & 0.8881 & 0.77\% \\
& & \textbf{CAT}  & 0.9527 & 0.9515 & 0.9467 & 0.9461 & 0.9469 & \textbf{0.69\%} & 0.9521 & 0.9507 & 0.9503 & 0.9499 & \textbf{0.29\%} \\
\midrule

\multirow{6}{*}{AUC}
& \multirow{3}{*}{\ding{172}} 
& Medley        & 0.9440 & 0.9399 & 0.9431 & 0.9350 & 0.9239 & 2.13\% & 0.9428 & 0.9347 & 0.9384 & 0.9364 & 0.99\% \\
& & TrustGuard    & 0.9382 & 0.9340 & 0.9342 & 0.9249 & 0.9210 & 1.83\% & 0.9338 & 0.9329 & 0.9341 & 0.9322 & 0.64\% \\
& & \textbf{CAT}  & 0.9677 & 0.9689 & 0.9673 & 0.9683 & 0.9681 & \textbf{0.04\%} & 0.9695 & 0.9663 & 0.9648 & 0.9660 & \textbf{0.30\%} \\
\cmidrule{2-14}
& \multirow{3}{*}{\ding{173}} 
& Medley        & 0.7806 & 0.7474 & 0.6941 & 0.6820 & 0.6596 & 15.50\% & 0.7636 & 0.7570 & 0.7619 & 0.7383 & 5.42\% \\
& & TrustGuard    & 0.7312 & 0.7252 & 0.7261 & 0.7079 & 0.7030 & 3.86\% & 0.7218 & 0.7216 & 0.7234 & 0.7216 & 1.31\% \\
& & \textbf{CAT}  & 0.8933 & 0.8898 & 0.8765 & 0.8748 & 0.8776 & \textbf{2.07\%} & 0.8913 & 0.8867 & 0.8875 & 0.8870 & \textbf{0.74\%} \\

\bottomrule[1.5pt]
\end{tabular}

\end{table*}

\section{Details of Evaluation Metrics} \label{metrics}
For clarity, we define a positive link as a trusted relationship between a trustor-trustee pair and a negative link as a distrusted relationship in the following metrics:
\begin{itemize}
\item\textbf{MRR:} MRR is the mean of the reciprocal rank over all positive links. 
\begin{equation}
    \text{MRR} = \frac{1}{N} \sum_{i=1}^{N} \frac{1}{\text{rank}_i},
    \nonumber
\end{equation}
where $N$ is the total number of positive links, and $\text{rank}_i$ is the rank of the $i$-th positive link according to its prediction score. To calculate MRR, we randomly sample 100 negative links for each positive link and rank the positive link among these negatives.

\item\textbf{AP:} AP summarizes a precision-recall curve that shows the trade-off between precision and recall for different classification thresholds.
\begin{equation}
    \text{AP} = \sum_{n}^{} (R_n - R_{n-1})P_n,
    \nonumber
\end{equation}
where $P_n$ and $R_n$ are the precision and recall at the $n$-th classification threshold.

\item\textbf{AUC:} AUC measures the likelihood that a positive link is ranked higher than a random negative link.
\begin{equation}
    \text{AUC} = \frac{\textstyle \sum_{i \in \mathcal{S}^1}^{} \sum_{j \in \mathcal{S}^2}^{} \mathbf{1}(\text{rank}_i < \text{rank}_j)}{|\mathcal{S}^1| \cdot |\mathcal{S}^2|},
    \nonumber
\end{equation}
where $\mathcal{S}^1$ and $\mathcal{S}^2$ are sets of positive and negative links, respectively. The indicator function $\mathbf{1}$ outputs 1 if $\text{rank}_i < \text{rank}_j$, and 0 otherwise.
\end{itemize}

\section{Robustness Evaluation on Trust-oriented and GNN-oriented Attacks} \label{attack_setting}
\textbf{Attack Settings.} For GNN-oriented data poisoning attacks, to the best of our knowledge, T-Spear~\cite{lee2024spear} is the only recent method targeting continuous-time GNNs. It formulates the attack by considering four constraints to achieve unnoticeability, which has proven effective in several popular continuous-time GNNs. Following the setting of T-Spear, we inject different numbers of adversarial links (i.e., malicious trust relationships) into the training and validation stages by adjusting perturbation rates. After training with poisoned data, we calculate metrics based on the ground truth trust relationships in the test set.

For trust-oriented data poisoning attacks, we follow the setting of TrustGuard~\cite{trustguard} to inject different proportions of malicious trust relationships. Since TrustGuard operates as a discrete-time model that does not provide timestamps for injected links, we generate timestamps by sampling from the temporal distribution of the original links. This approach, consistent with T-Spear~\cite{lee2024spear}, helps maintain attack stealthiness.

\textbf{Results.} Fig.~\ref{robustness_apauc} presents the AP and AUC results of CAT and baseline models under two types of attacks. CAT exhibits the highest robustness, with a maximum performance drop of only 2.07\%. In contrast, Medley shows limited robustness due to its lack of defense mechanisms, especially in task~\ding{173}. Notably, all models experience minor performance drops in most cases. The reason is that both attacks simulate realistic threat scenarios, where attackers introduce some constraints to make their attacks stealthy and unnoticeable, thus limiting their attack strength.

\begin{table*}[htbp]
\footnotesize
\centering
\caption{Robustness comparison under the adaptive attack (MRR results).}
{\vspace{-1mm}\ding{172}/\ding{173}: trust predictions for observed/unobserved users. $p$: perturbation rate\\(ratio of added adversarial links to original links). MDR: Maximum Drop Rate (lower is better).}
\\[2mm]
\label{adaptive_attack}
\begin{tabular}{c|c|ccccc|ccccc}
\toprule[1.5pt]
\multirow{2.5}{*}{Tasks} & \multirow{2.5}{*}{Clean}
& \multicolumn{5}{c|}{\textbf{CAT}} & \multicolumn{5}{c}{w/o Context-aware Aggregation} \\
\cmidrule{3-7} \cmidrule{8-12}
& & $p$=5\% & $p$=10\% & $p$=15\% & $p$=20\% & MDR $\downarrow$
  & $p$=5\% & $p$=10\% & $p$=15\% & $p$=20\% & MDR $\downarrow$ \\
\midrule
\ding{172} & 0.6025 & 0.5855 & 0.5801 & 0.6025 & 0.5878 & \textbf{3.72\%} & 0.5748 & 0.5335 & 0.5654 & 0.5569 & 11.45\% \\
\ding{173} & 0.4082 & 0.3885 & 0.3754 & 0.3774 & 0.3742 & \textbf{8.33\%} & 0.3841 & 0.3683 & 0.3763 & 0.3713 & 9.77\% \\
\bottomrule[1.5pt]
\end{tabular}
\vspace{-3mm}
\end{table*}

\section{Robustness Evaluation on Adaptive Attacks} \label{appendix_adaptive}
\textbf{Attack Settings.} The adaptive attack, referred to as the cross-context bridging attack, aims to blur the boundaries between two distinct contexts by injecting malicious user-item interactions, causing CAT to confuse them. The attack setting is as follows: (i) \textit{Context selection:} We randomly select two distinct item categories from the Epinions dataset as contexts $c_k$ and $c_{k'}$. (ii) \textit{Colluders selection:} We choose highly-active users with substantial interactions in $c_k$ as colluders. These users are chosen because they capture the most representative behavioral patterns of $c_k$. Thus, when they engage extensively in $c_{k'}$, their behavior introduces misleading cross-context signals that make CAT confuse distinct contexts as similar, thereby distorting its learned contextual boundaries. (3) \textit{Attack injection:} Each colluder injects new interactions with randomly chosen items in $c_{k'}$. The timestamps of these injected interactions are uniformly distributed across the dataset's historical time span to mimic normal temporal patterns and avoid detection. The total number of injected links is controlled by a perturbation rate, and a balanced allocation strategy specifies both the number of colluders and the number of links injected by each colluder.

We evaluate CAT's robustness against this strong attack under perturbation rates of 5\%, 10\%, 15\%, and 20\%. Since baseline models do not incorporate contextual information, the attack is not applicable to them. Instead, we construct a variant by removing CAT's context-aware aggregation mechanism to assess its contribution to model robustness. We present only MRR results, as AUC and AP exhibit similar trends.

\textbf{Results.} As shown in Table~\ref{adaptive_attack}, CAT experiences a maximum performance drop of 3.72\% in task \ding{172}, compared with 0.95\% under trust-oriented attacks and 3.39\% under GNN-oriented attacks. For task \ding{173}, the drop reaches 8.33\%, higher than 4.63\% and 5.49\% under the two respective attacks. These results indicate that the cross-context bridging attack is more effective than trust-oriented and GNN-oriented attacks, as it directly targets CAT's context-aware design. Nevertheless, CAT remains robust in task \ding{172}, while its performance degrades more noticeably in task \ding{173}. This is because the attack makes the semantic boundaries between contexts less sharp, making it more difficult for CAT to learn distinguishable patterns that generalize to unobserved users.

When the context-aware aggregation mechanism is removed, CAT's robustness declines substantially, with the maximum performance drops increasing to 11.45\% in task~\ding{172} and 9.77\% in task~\ding{173}. This confirms the critical role of context-aware aggregation in enhancing model robustness, as it mitigates noise and preserves contextual distinctions by adaptively weighting contexts.

\section{Comparison of Two Neighbor Sampling Strategies}
\label{appendix_sampling}
CAT adopts a recent-time neighbor sampling strategy to enhance scalability. In this section, we compare this strategy with the uniform neighbor sampling strategy, which uniformly samples nodes and is commonly used in static graph analysis~\cite{graphsage}. The results in Table~\ref{comparsion_sampling} are based on the Epinions dataset using a 70\%-15\%-15\% split ratio. From the table, we can see that recent-time neighbor sampling is more effective and efficient than uniform neighbor sampling, indicating that recent interactions are more informative and important than historical ones in dynamic graphs.

\begin{table}[tbp]
\footnotesize
\centering
\caption{Comparison of uniform and recent-time neighbor sampling strategies.}
\label{comparsion_sampling}
{\vspace{-1mm}\ding{172}/\ding{173}: trust predictions for observed/unobserved users.}
\\[2mm]
\begin{tabular}{c|c|c|c|c|c}
\toprule[1.5pt]
Strategies     & Tasks    & MRR    & AP    & AUC   & \makecell[c]{Time\\(s/epoch)}   \\ \midrule
\multirow{2}{*}{Uniform}      
        & \ding{172} & 0.5907 & 0.9368 & \textbf{0.9678} & \multirow{2}{*}{52.73} \\
        & \ding{173} & 0.3735 & 0.9487 & 0.8861 &                       \\ \midrule
\multirow{2}{*}{\textbf{Recent-time (CAT)}} 
        & \ding{172} & \textbf{0.6025} & \textbf{0.9383} & 0.9677 & \multirow{2}{*}{\textbf{48.04}} \\
        & \ding{173} & \textbf{0.4082} & \textbf{0.9527} & \textbf{0.8933} &                       \\
\bottomrule[1.5pt]
\end{tabular}
\end{table}

\section{Comparison of Context Embedding Generators} \label{context_generator}
In addition to the average generator used in CAT, we develop two alternative generators for context embeddings: (i) Long Short-Term Memory (LSTM) generator: An LSTM is used to generate a context embedding by processing an unordered set of item embeddings within the same context. (ii) MLP generator: An MLP aggregates item embeddings with the same context, followed by an elementwise max-pooling operation to obtain a context embedding. Results in Table~\ref{comparison_generator} show that our simple, efficient average generator outperforms both alternatives. One possible reason is that the LSTM and MLP architectures introduce unnecessary complexity, increasing the risk of overfitting.

\begin{table}[tbp]
\footnotesize
\centering
\caption{Comparison of different generators for context embeddings.}
\label{comparison_generator}
{\vspace{-1mm}\ding{172}/\ding{173}: trust predictions for observed/unobserved users.}
\\[2mm]
\begin{tabular}{c|c|c|c|c|c}
\toprule[1.5pt]
Generators     & Tasks    & MRR    & AP    & AUC   & \makecell[c]{Time\\(s/epoch)}   \\ \midrule
\multirow{2}{*}{LSTM}      
        & \ding{172} & 0.5191 & 0.9168 & 0.9608 & \multirow{2}{*}{405.08} \\
        & \ding{173} & 0.3510 & 0.9458 & 0.8824 &                       \\ 
\midrule
\multirow{2}{*}{MLP} 
        & \ding{172} & 0.4794 & 0.9091 & 0.9586 & \multirow{2}{*}{77.47} \\
        & \ding{173} & 0.3255 & 0.9448 & 0.8841 &                       \\ 
\midrule
\multirow{2}{*}{\textbf{Avg (CAT)}} 
        & \ding{172} & \textbf{0.6025} & \textbf{0.9383} & \textbf{0.9677} & \multirow{2}{*}{\textbf{48.04}} \\
        & \ding{173} & \textbf{0.4082} & \textbf{0.9527} & \textbf{0.8933} &            \\
\bottomrule[1.5pt]
\end{tabular}
\vspace{-3mm}
\end{table}

\section{Visualization of Dual Attention Scores} \label{visualization_dual}
CAT comprises several components, making it difficult to understand how it works and whether its functionality aligns with human expectations. To address this, we visualize two key components of CAT: the dual attention mechanism and the context-awareness aggregation mechanism, to explain its prediction process. Since the latter was investigated in Section~\ref{rq1}, this section focuses on the dual attention mechanism.

\begin{figure}[tbp]
	\centering
	\includegraphics[scale=0.52]{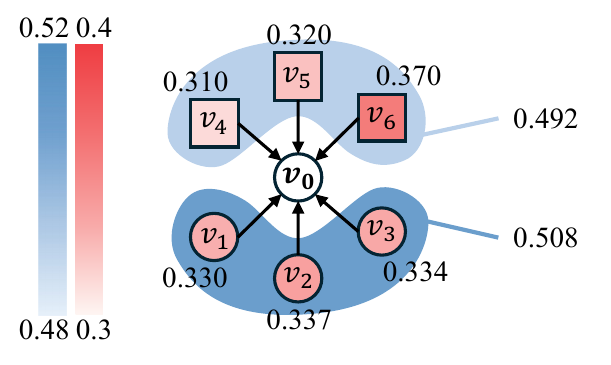}
	\caption{Visualization of the dual attention scores, including type attention scores (shown in blue) and node attention scores (shown in red). $v_0 \sim v_3$ denote users and $v_4 \sim v_6$ denote items, with $v_0$ being the target node.}
	\label{attention_example}
\end{figure}

As shown in Fig.~\ref{attention_example}, $v_0$ represents the target user who interacts with both items and other users. We observe that different weights are assigned to the user type (i.e., $\langle user, user \rangle$ interactions) and the item type (i.e., $\langle item, user \rangle$ interactions), with the user type receiving higher importance. This is reasonable since our task is trust prediction, where trust relationships between user pairs directly affect the outcomes. Additionally, the node attention scores vary across nodes, and this variation is particularly evident within the item type. For example, the importance of $v_6$ to $v_0$ is 0.37, whereas the importance of $v_4$ to $v_0$ is only 0.31. Overall, this visualization suggests that the dual attention mechanism can effectively handle heterogeneous graphs by distinguishing both interaction types and individual nodes. However, the granularity of explanations could be further improved by investigating specific explanation methods, such as T-GNNExplainer~\cite{xia2023explaining}, which we leave as future work. Such methods not only improve user trust in the model but also facilitate its practical adoption~\cite{Han_Explanation}.

\begin{figure}[tbp]
    \centering
    \subfigcapskip=-3pt

    \subfigure[CAT]{\includegraphics[scale=0.5]{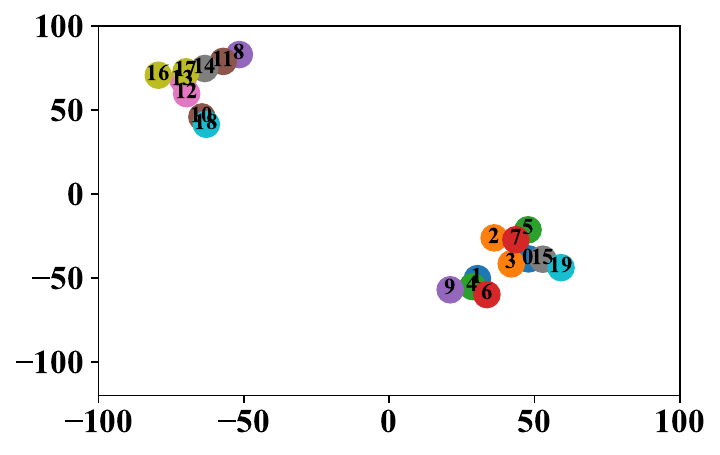}}

    \vspace{-3mm}
    
    \subfigure[HGT]{\includegraphics[scale=0.5]{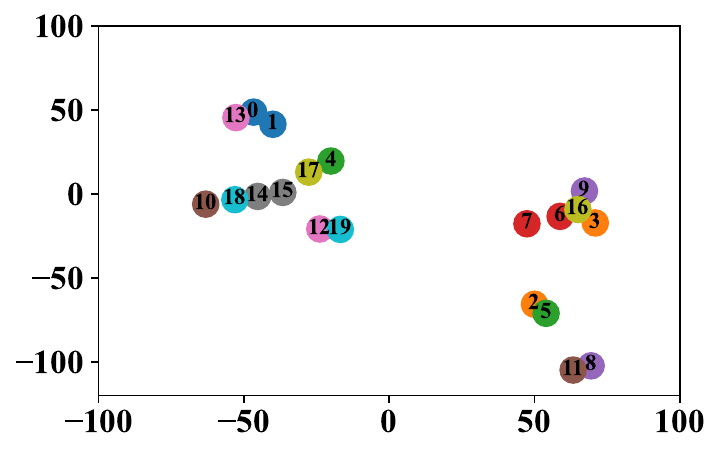}}
    
    \vspace{-1mm}
    \caption{Two-dimensional t-SNE visualization of user pairs with trusted relationships, where each pair is shown in the same color.}
    \label{visualization_emb}
    \vspace{-2mm}
\end{figure}

\section{Visualization of Node Embeddings} \label{appendix_emb}
To evaluate the quality of node embeddings learned by different trust prediction models, we visualize user embeddings for trusted pairs using the t-SNE technique~\cite{van2008visualizing}. For clarity, we randomly select 10 user pairs with trusted relationships from the Epinions dataset, as illustrated in Fig.~\ref{visualization_emb}. We choose HGT for comparison as it performs the best among the baseline models. Intuitively, when two users have a trusted relationship, they should be located close to each other in the embedding space~\cite{trustguard,cdeeptrust}. In the figure, this means nodes of the same color should be as close as possible. We observe that CAT produces generally shorter distances between trusted pairs compared to HGT, indicating that CAT effectively captures complex trust relationships in dynamic heterogeneous networks.

\end{document}